\documentclass[twocolumn]{svjour3}          

\smartqed  

\usepackage{cite}
\usepackage{amsmath,amssymb,amsfonts}
\usepackage{algpseudocode,algorithm}
\usepackage{graphicx}
\usepackage{textcomp}

\begin{document}

\sloppy

\title{Sample-efficient Gear-ratio Optimization for Biomechanical Energy Harvester
\thanks{This work was supported by JSPS KAKENHI, Grant Number JP16H05876.}
}
\subtitle{}

\titlerunning{Gear-ratio Optimization for Energy Harvester}        

\author{Taisuke Kobayashi \and Yutaro Ikawa \and Takamitsu Matsubara}


\institute{T. Kobayashi, Y. Ikawa, and T. Matsubara \at
              Division of Information Science, Graduate School of Science and Technology, Nara Institute of Science and Technology, Nara 630-0192, Japan.\\
              \email{kobayashi@is.naist.jp}           
}

\date{Received: date / Accepted: date}

\maketitle
\begin{abstract}

The biomechanical energy harvester is expected to harvest the electric energies from human motions.
A tradeoff between harvesting energy and keeping the user's natural movements should be balanced via optimization techniques.
In previous studies, the hardware itself has been specialized in advance for a single task like walking with constant speed on a flat.
A key ingredient is Continuous Variable Transmission (CVT) to extend it applicable for multiple tasks.
CVT could continuously adjust its gear ratio to balance the tradeoff for each task; however, such gear-ratio optimization problem remains open yet since its optimal solution may depend on the user, motion, and environment.
Therefore, this paper focuses on a framework for data-driven optimization of a gear ratio in a CVT-equipped biomechanical energy harvester.
Since the data collection requires a heavy burden on the user, we have to optimize the gear ratio for each task in the shortest possible time.
To this end, our framework is designed sample-efficiently based on the fact that the user encounters multiple tasks, which are with similarities with each other.
Specifically, our framework employs multi-task Bayesian optimization to reuse the optimization results of the similar tasks previously optimized by finding their similarities.
Through experiments, we confirmed that, for each task, the proposed framework could achieve the optimal gear ratio of around 50~\% faster than one by random search, and that takes only around 20~minutes.
Experimental results also suggested that the optimization can be accelerated by actively exploiting similarities with previously optimized tasks.

\end{abstract}

\section{Introduction}
\label{sec:introduction}

In the Internet of Things (IoT) society, most people carry a lot of electronic devices like a smartphone(s), tablet(s), laptop pc(s), and so on.
These devices are usually battery-driven, but their capacity is insufficient to use all day due to size limitations.
Therefore, the users are often forced to carry mobile (or extra) batteries together.
While the batteries store electric energy directly, humans can store chemical energy from food products.
Indeed, the energy efficiency that humans can store and output is reported to be 35--100 times higher than that of the most currently available batteries~\cite{rome2005generating}.
In other words, humans' daily activities consume roughly $1.07 \times 10^7$~J, which is comparable to the electric energy stored in a 15~kg battery~\cite{riemer2011biomechanical,schertzer2015harvesting}.
Even if limited to walking, it is theoretically able to harvest 5~J energy from the landing of walking, and 900~MWh per day can be expected to be generated in the UK~\cite{partridge2016analysis}.
In this way, humans can be regarded as excellent energy sources instead of a battery if energy can easily be gained from us.

The concept of \textit{energy harvesting}, which generates energy from human motions, has been spread over the world in the last decade~\cite{rome2005generating,donelan2008biomechanical,shepertycky2015generating}.
The research on \textit{energy harvester} is getting active to solve the open issue of energy shortage at the individual level, although no practical device has been developed yet.
The previous studies mainly focus on energy harvesting from walking since it is an essential daily motion.
In particular, they alternate the negative work like braking knee extensor just before landing (i.e., at swing-leg retraction)~\cite{devita2007muscles,winter2009biomechanics}.
In that case, the energy harvester's reaction force would enable the highly efficient harvesting with reduced human loads while keeping the natural movements.
Note that this paper defines the natural movements as the movements without any additional efforts and reduced negative work as much as possible.
The human biological signal has been processed to judge the negative work in walking~\cite{selinger2015myoelectric}.
The optimal gearbox design has also been carried out to optimize both the alternative negative work and the energy harvesting at that time~\cite{jhalani2012optimal}.

Commonly for all the previous studies above, the tradeoff between the harvesting energy capacity and keeping the natural human movements needs to be suitably balanced at hardware design (e.g., \cite{jhalani2012optimal}).
Such an optimal design can reflect the physical differences between users.
However, even with the design well-fitted for each user, the previous energy harvesters cannot adapt to various environments and motions (let's call \textit{tasks}) since they have not the capability to balance the above tradeoff.
Therefore, the previous studies focused on a single task like walking with constant speed on a flat.

\begin{figure}[tb]
    \centering
    \includegraphics[keepaspectratio=true,width=0.9\linewidth]{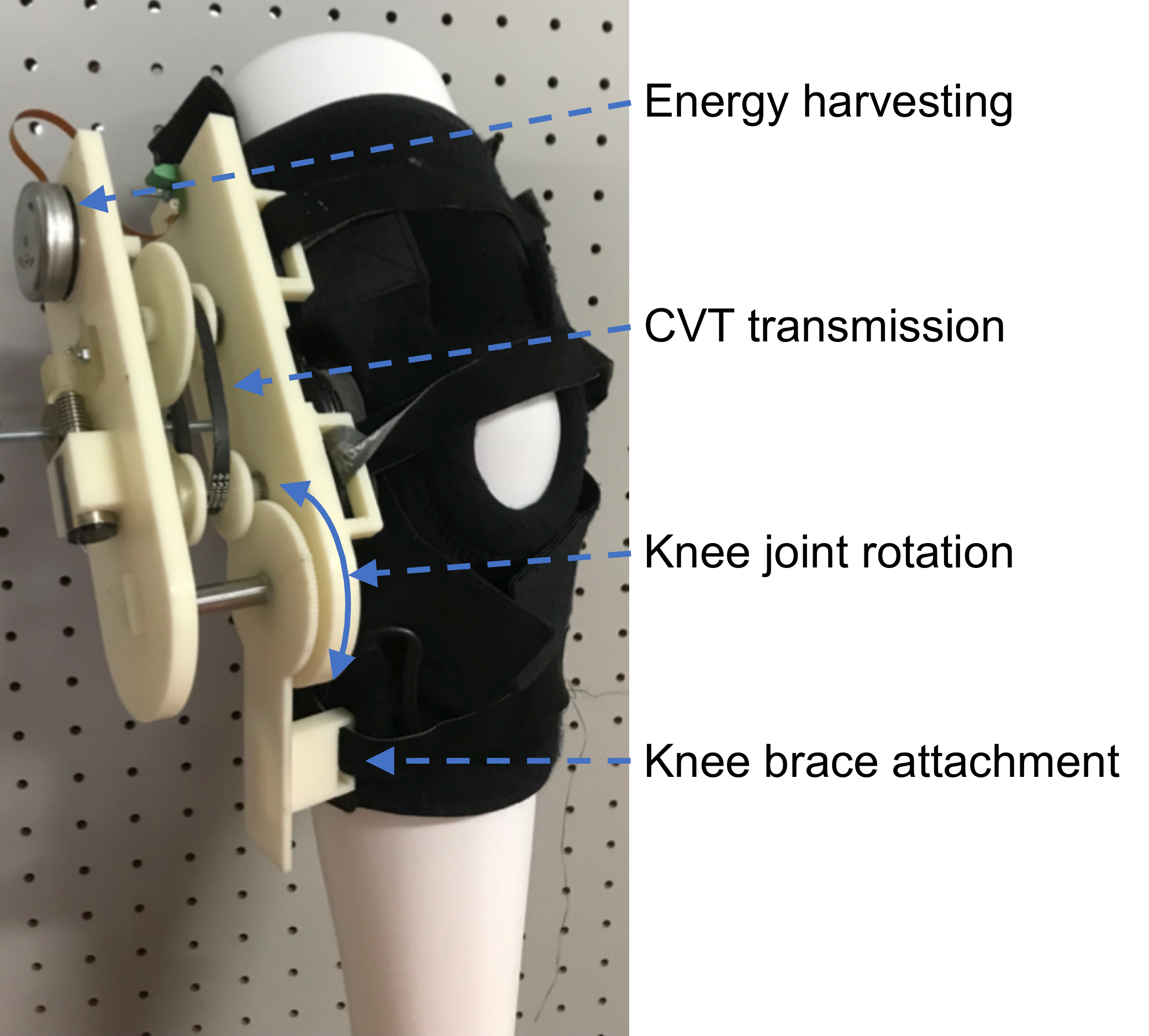}
    \caption{CVT-equipped energy harvester for a knee joint:
    this device is attached to the outside of a knee using a knee brace;
    a knee joint rotation is transmitted to an electromagnetic generator through CVT;
    the gear ratio can be changed continuously by rotating a screw via a small motor with a worm gear.
    }
    \label{fig:EH_promotion}
\end{figure}

To apply the energy harvester to multiple tasks, Continuous Variable Transmission (CVT)-equipped biomechanical energy harvester has been developed (see Fig.~\ref{fig:EH_promotion})~\cite{ikawa2018biomechanical, singla2016optimization}.
By adjusting the gear ratio in CVT, this device could balance the harvesting energy capacity and keep the natural human movements.
Such a well-designed hardware device allows us to formulate the tradeoff problem as a control-parameter optimization problem for balancing the tradeoff.

However, such a gear-ratio optimization problem remains open yet since its optimal solution may depend on the user, motion, and environment.
In other words. the device should be optimized while being attached to the user among various tasks, although the user cannot stand the numerous trials for optimization.
The recent deep learning methods~\cite{lecun2015deep} and reinforcement learning methods~\cite{sutton2018reinforcement} are not suitable for this optimization problem since they require numerous samples collected through interaction between the energy harvester and the user equipped with device.
We have to resolve this optimization problem from a few trials.
The system then finds the optimal gear ratio while reducing the user effort required for daily life with the energy harvester.

In this paper, we propose a framework for data-driven sample-efficient optimization of a gear ratio in a CVT-equipped biomechanical energy harvester.
To this end, we raise two reasonable assumptions as follows:
\begin{enumerate}
\item The gear ratio contributes to balancing the tradeoff according to a specific rule (function).
\item The various tasks are not independent of each other, and their similarities can be identified.
\end{enumerate}
With these assumptions, our framework employs multi-task Bayesian optimization~\cite{swersky2013multi} for the core of the optimization process.
Bayesian optimization~\cite{srinivas2010gaussian} can reveal the task-specific balance function w.r.t the gear ratio in an efficient way and can choose the optimal gear ratio according to the revealed function.
Our framework further enhances its sample-efficiency for achieving new tasks by transferring previous optimization results for different tasks according to similarities between them.
This efficient adaptability to new tasks would enable our device to harvest energy from natural motion for multiple tasks (e.g., different slope angles and gait speeds).

We conducted real-world experiments with the developed energy harvester attached to two participants.
The similarities between tasks were adequately revealed in our approach.
As a result, experimental results suggest that our approach can optimize task-specific gear ratios from fewer trials than a random search.
Indeed, although the random search requires around 60~minutes for the optimization per task, our framework takes only around 20~minutes.
When the new task is similar to the tasks previously optimized, that time can be reduced more.

Our contributions are two folds.
The first one is to develop a data-driven sample-efficient framework that can optimize gear ratios in CVT-equipped energy harvester with the user over multiple tasks based on multi-task Bayesian optimization.
The second is to verify the sample-efficiency of our proposed framework in real-world experiments.
Namely, our contributions would be an essential step in shifting the old way of energy harvesters for simple tasks toward daily tasks.
Note that our contributions in the above are on the general framework for the gear ratio optimization to handle the tradeoff, not on the specific optimal balance of the tradeoff and maximization of the amount of harvesting energy.
Although the importance of optimal hardware design is clear from previous studies (e.g., \cite{jhalani2012optimal}),
this paper focuses only on gear ratio optimization; namely, the primary purpose is to show adaptability to multiple gear ratio optimization tasks.

The remainder of this paper is structured as follows.
Section~\ref{sec:development} gives a brief introduction of a CVT-equipped energy harvester we developed, and why the gear ratio is suitable for being optimized is also illustrated.
Section~\ref{sec:optimization} shows our human-in-the-loop optimization framework for the gear ratio in CVT through a few trial-and-error.
Section~\ref{sec:setup} sets our experimental conditions with human participants.
Section~\ref{sec:result} shows and analyzes the experimental results.
Section~\ref{sec:discussion} discusses limitations of the proposed framework and suggests solutions for them.
Section~\ref{sec:conclusion} gives a summary and potential future work.

\section{CVT-equipped Energy Harvester}
\label{sec:development}
%
\subsection{Principle of energy harvesting}
\label{subsec:principle}

\begin{figure*}[tb]
    \centering
    \includegraphics[keepaspectratio=true,width=0.9\linewidth]{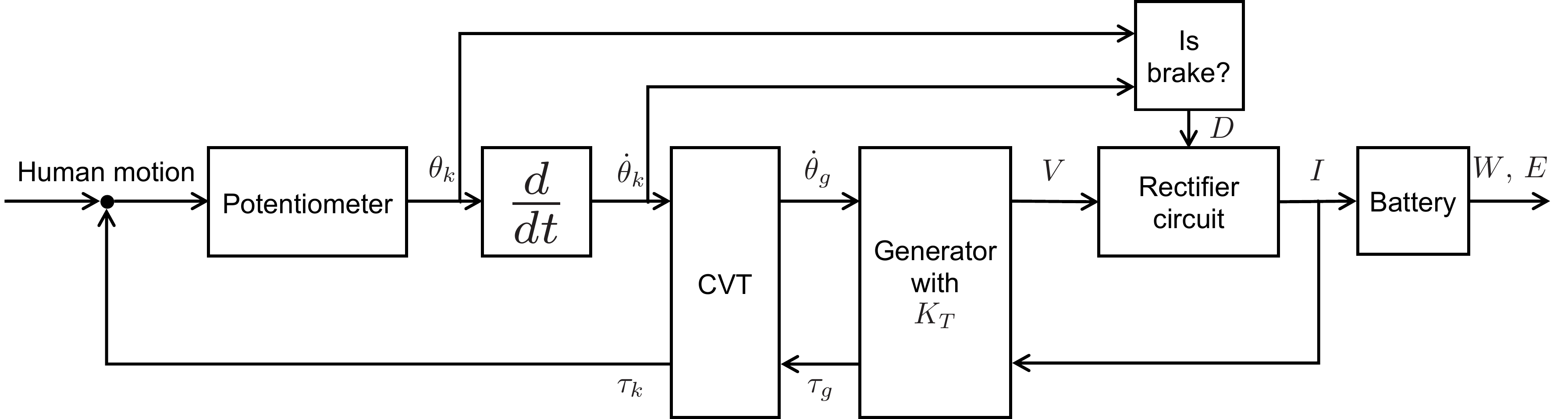}
    \caption{Flow diagram of energy harvesting:
    the human motion is measured by a potentiometer;
    the knee joint motion is transmitted to the generator and converted to the generative power;
    on the other hand, the generative power (current) is converted to the anti torque to the knee joint.
    }
    \label{fig:EH_flow}
\end{figure*}

First, to show the effectiveness of CVT, the generative power $W$ and the anti-torque to the knee $\tau_k$ are derived theoretically.
A flow diagram of our device is depicted in Fig.~\ref{fig:EH_flow}.

\subsubsection{Derivation of the generative power and the anti torque}

The angular velocity of the knee joint $\dot{\theta}_k$, which can be measured by a potentiometer, is amplified by the gear ratio, $G$, to the angular velocity of the generator $\dot{\theta}_g$.
The counter-electromotive force of the generator, $V$, is proportional to $\dot{\theta}_g$ with $K_T$ the torque constant of the generator.
When the generator is connected to a rectifier circuit with the duty cycle $D \in [0, 1]$, the current in the circuit, $I$, is derived as follows:
\begin{align}
    I = D \frac{V}{R} = \frac{K_T G D}{R} \dot{\theta}_k
    \label{eq:V2I}
\end{align}
where $R$ is the internal resistance of the circuit.
Finally, $W$ is given from Ohm's law.
\begin{align}
    W = R I^2 = \frac{K_T^2 G^2 D^2}{R} \dot{\theta}_k^2
    \label{eq:watt}
\end{align}

On the other hand, the anti-torque to the generator, $\tau_g$, is the product of $K_T$ and $I$.
The gear ratio from the generator to the knee is given as $1/G$, and the torque is inversely proportional to the gear ratio.
That is, $\tau_k$ is derived as follows:
\begin{align}
    \tau_k = G \tau_g = G K_T I = \frac{K^2_T G^2 D}{R} \dot{\theta}_k
    \label{eq:torque}
\end{align}

\subsubsection{Control parameter for resolving tradeoff}

As can be seen in eqs.~\eqref{eq:watt} and \eqref{eq:torque}, both $W$ and $\tau_k$ show qualitatively similar behaviors.
That means, a tradeoff between harvesting energy and reducing the anti-torque, which would keep the user's movements natural, would be expected.
Here, $K_T$ and $R$ are constant in the developed system, although they should be large and small as much as possible, respectively, for energy efficiency.
In addition, $\dot{\theta}_k$ is motion-dependent.
If $G \in [G_\mathrm{min}, G_\mathrm{max}]$ and $D \in [0, 1]$ can be tuned within their ranges by the CVT and a microcontroller, both $W$ and $\tau_k$ will be optimizable.

The difference between them is the magnitude of their effects: $W$ and $\tau_k$ are the functions w.r.t $G^2$ although they are the functions w.r.t $D^2$ and $D$, respectively.
That is, the gear ratio $G$ is suitable for weakening $\tau_k$ while keeping $W$.
We therefore choose $G$ as an control parameter to be optimized, although its adjustment would take a longer time than that of $D$.
Reducing the cost during the optimization should be resolved for practical use.

\subsection{Hardware design}
\label{subsec:hardware}

\begin{figure}[tb]
    \centering
    \includegraphics[keepaspectratio=true,width=0.95\linewidth]{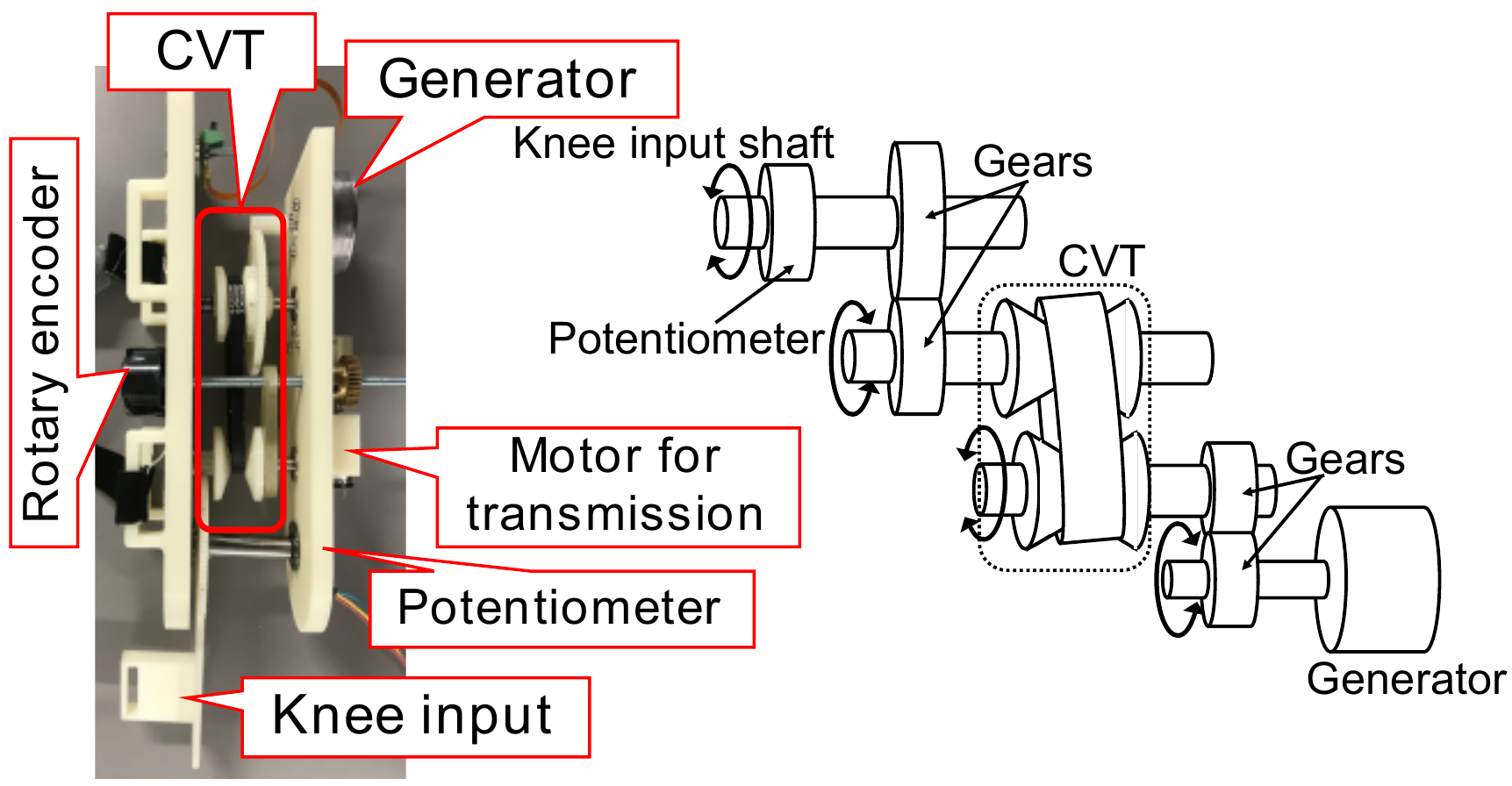}
    \caption{Mechatronics of the developed energy harvester:
    from a knee input, the user's knee motion is transmitted to a electromagnetic generator through a belt-type CVT, which can be tuned by rotating a motor for transmission via a worm gear.
    }
    \label{fig:EH_structure}
\end{figure}

\begin{table}[tb]
    \centering
    \small
    \caption{Specification of mechanical parameters}
    \begin{tabular}{ c c r r }
        \hline
        Symbol& Meaning & Value &  Unit
        \\ \hline
        $m$ & Mass & 0.78 & kg
        \\
        $\theta_k$ & Movable range of knee & [0, $\pi/2$] & rad
        \\
        $G$ & Gear ratio & [16, 144] & ---
        \\
        $K_T$ & Torque constant & $70.6 \times 10^{-3}$ & Nm/A
        \\
        $R$ & Internal resistance & $1.4$ & $\mathrm{\Omega}$
        \\ \hline
    \end{tabular}
    \label{tab:params_mech}
\end{table}

To adjust the gear ratio for the effective optimization of $W$ and $\tau_k$, a new energy harvester with CVT has been developed in our previous work~\cite{ikawa2018biomechanical}.
Note that the minor improvements from the prototype~\cite{ikawa2018biomechanical} were conducted in the magnitude of generative power by changing a generator (i.e., $K_T$) and a range of the gear ratio $[G_\mathrm{min}, G_\mathrm{max}]$ while keeping the device light-weight.

Let us introduce its mechatronics as below.

\subsubsection{Overview}

Its mechanical structure is shown in Fig.~\ref{fig:EH_structure}, and mechanical parameters are in Table~\ref{tab:params_mech}.
A knee brace (McDavid: M429X) is used to attach the device to the user.
A link is rotated according to the knee motion, and this rotation is transmitted to a electromagnetic generator via three transmissions, one of which is a belt-type CVT.
Note that the rigid parts except the shafts are designed with a 3D printer, materials of which are ABS plus-P430, to reduce its weight.
The developed energy harvester is light-weight compared to one developed in the previous work~\cite{donelan2008biomechanical}, and has the movable range of knee sufficient for walking and jogging.

\subsubsection{CVT mechanism}

\begin{figure}[tb]
    \centering
    \includegraphics[keepaspectratio=true,width=0.95\linewidth]{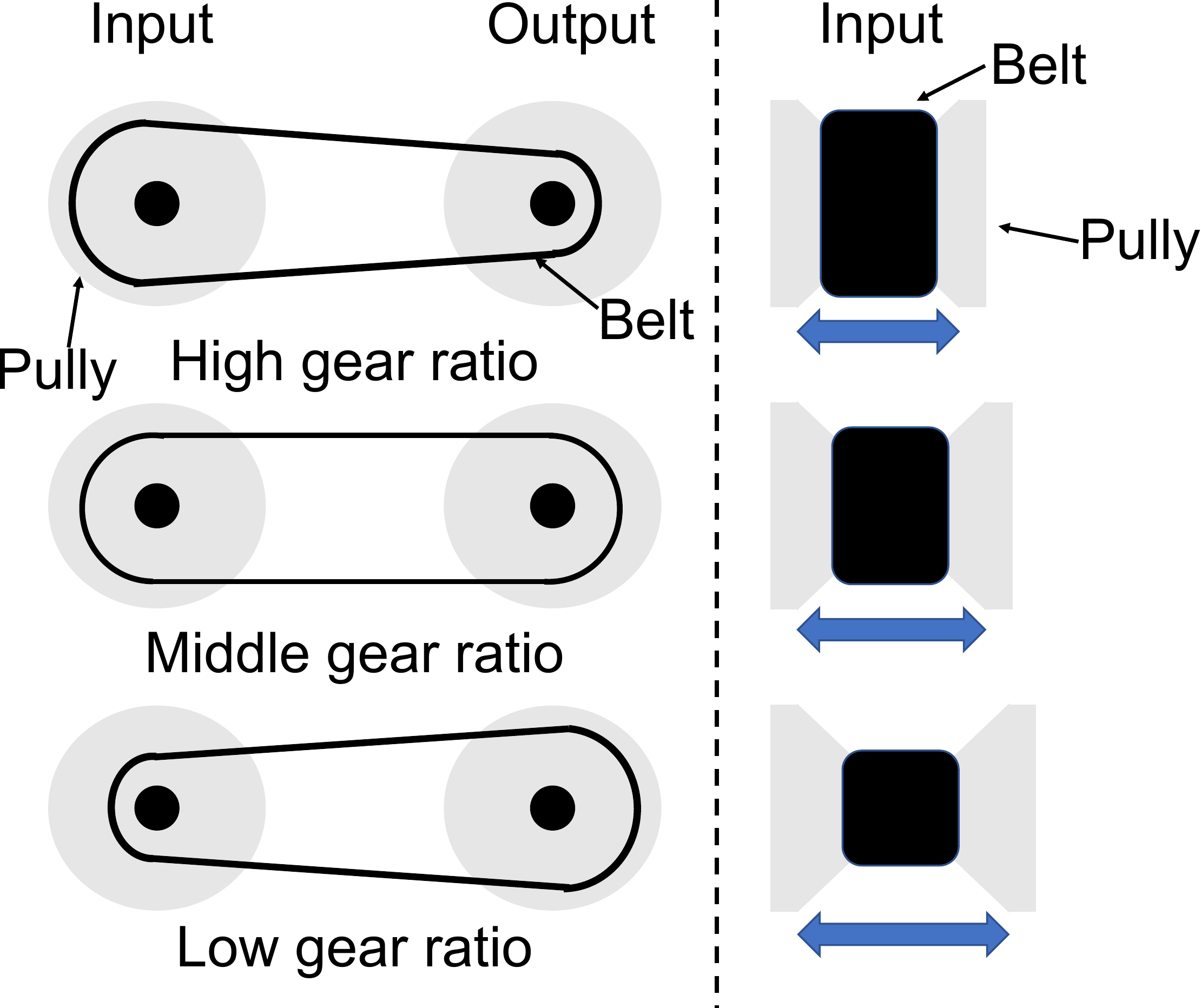}
    \caption{Illustration of a belt-type CVT:
    pulleys on both side can change the distance between their cones;
    a rubber belt adapts to the pulleys' motion by its tension, and therefore, the effective diameters are changed;
    using the rubber's friction, the rotation of one side is transmitted to that of another side.
    }
    \label{fig:EH_CVT}
\end{figure}

As mentioned in the above, a belt-type CVT~\cite{srivastava2009review}, which consists of two sets of variable-diameter pulleys, a rubber belt strung between them (see Fig.~\ref{fig:EH_CVT}), is installed in our energy harvester as one of the three transmissions.
The diameters of pulleys are changed when rotating a screw that they are attached by using a small motor (Portescap: 16N78- 208E.1001) via a worm gear.
Note that, due to no back-drivability of the worm gear, it enables to efficiently hold the diameters of the pulleys.
The gear ratio can be estimated from the rotation amount of the screw measured by a rotary encoder (Broadcom: HEDS-5540\#A06).
As a result, this CVT can tune its gear ratio from 1:3 to 3:1.
In total, with two fixed gear ratios (6:1 and 8:1, respectively), our energy harvester has the variable gear ratio within $G \in [G_\mathrm{min}, G_\mathrm{max}] = [16, 144]$.

\subsubsection{Electronics}

This harvester is equipped with a compact electromagnetic generator (Maxon Motor: EC45 flat), and rotational motion on a knee, which is measured by a potentiometer (Alps Alpine: RCD502010A), is transmitted to the generator via the transmissions.
When turning on the circuit that stores or measures the electric power from the generator, counter-electromotive force and anti-torque are simultaneously generated, as derived in the above.
Note that this turning on/off the circuit corresponds to the duty cycle $D$, and in our device, we decided to restrict it as binary $\{0,1\}$ for simplicity.
Here, the counter-electromotive force is rectified by a three-phase full-wave rectifier circuit.
This is because the knee joint reciprocates, not rotating in one direction, and the device is required to generate power from motions in both directions.
To choose when the device harvests energy (basically, it is better only when the user does negative work), rules to turn on the circuit can be considered kinetically~\cite{donelan2008biomechanical}, by electromyographic way~\cite{selinger2015myoelectric}, and so on.
For simplicity, this paper employs the rule depending on thresholds w.r.t the knee angle and angular velocity to exploit the anti-torque as negative work (i.e., braking motion)~\cite{ikawa2018biomechanical}.
%
\section{Proposed Framework for Data-driven Sample-efficient Gear-Ratio Optimization}
\label{sec:optimization}

\subsection{Overview}

We define our optimization problem mathematically and introduce the concept of our framework with Bayesian optimization for this problem.

\subsubsection{Problem statement}

Let $f: x \to y$ be the latent score function of the user with the harvester under some conditions, where input $x$ is the gear ratio, and output $y = f(x)$ is the score in terms of harvesting energy and keeping natural movements of the user, which is defined in the section~\ref{subsec:score}.
Here, we focus on the fact that this score is relevant to environments and motions (i.e., \textit{tasks}) when harvesting energy.
For example, walking or running on up or down slope would have the different tendency of the score.
That is, given $t$ as the symbol of task in addition to $x$, the task-specific score $f(x; t)$ is considered.
Our purpose is to find the optimal gear ratio $x^*$ that can maximize the score.
\begin{align}
x^* = \arg \max_{x \in \mathbb{R}^d} f(x; t)
\label{eq:opt_problem}
\end{align}
where $d$ is the dimension of the input ($d=1$ in our case).

The problem's difficulty is that the score function $f(x; t)$ is black-box.
In addition, sampling data from $f(x; t)$ is possible but expensive because it requires experiments in which the user wearing the harvester conducts motions on environments (specified by $t$) to obtain the score values according to the inputs $x$.
Thus, many samples cannot be collected; finding the optimum with the minimum amount of data and experiments is worth considering.

\subsubsection{Brief introduction of Bayesian optimization}

Bayesian optimization is a sequential design strategy for global optimization of black-box functions for quickly searching its global maxima.
Bayesian optimization has been successfully used for human-in-the-loop systems~\cite{matsubara2016data,thatte2017sample,gordon2018bayesian}; hence, it is expected to work well in the gear ratio optimization for our energy harvester.

In Bayesian optimization, an acquisition function, $\alpha (x; t)$, is alternatively optimized to find the next query point, $x^\prime$, as follows:
\begin{align}
x^\prime = \arg \max_{x \in \mathbb{R}^d} \alpha(x; t) \label{eq:sample_x}
\end{align}
Then, the query $x^\prime$ will be tested to obtain the new score $y^\prime$, and it is used to infer the score function more accurately and update the acquisition function.
These procedures are alternatively executed until the process finds the global optimal point $x^*$.
Using $\alpha$ yields efficient exploration.

\begin{figure}[tb]
    \centering
    \includegraphics[keepaspectratio=true,width=0.95\linewidth]{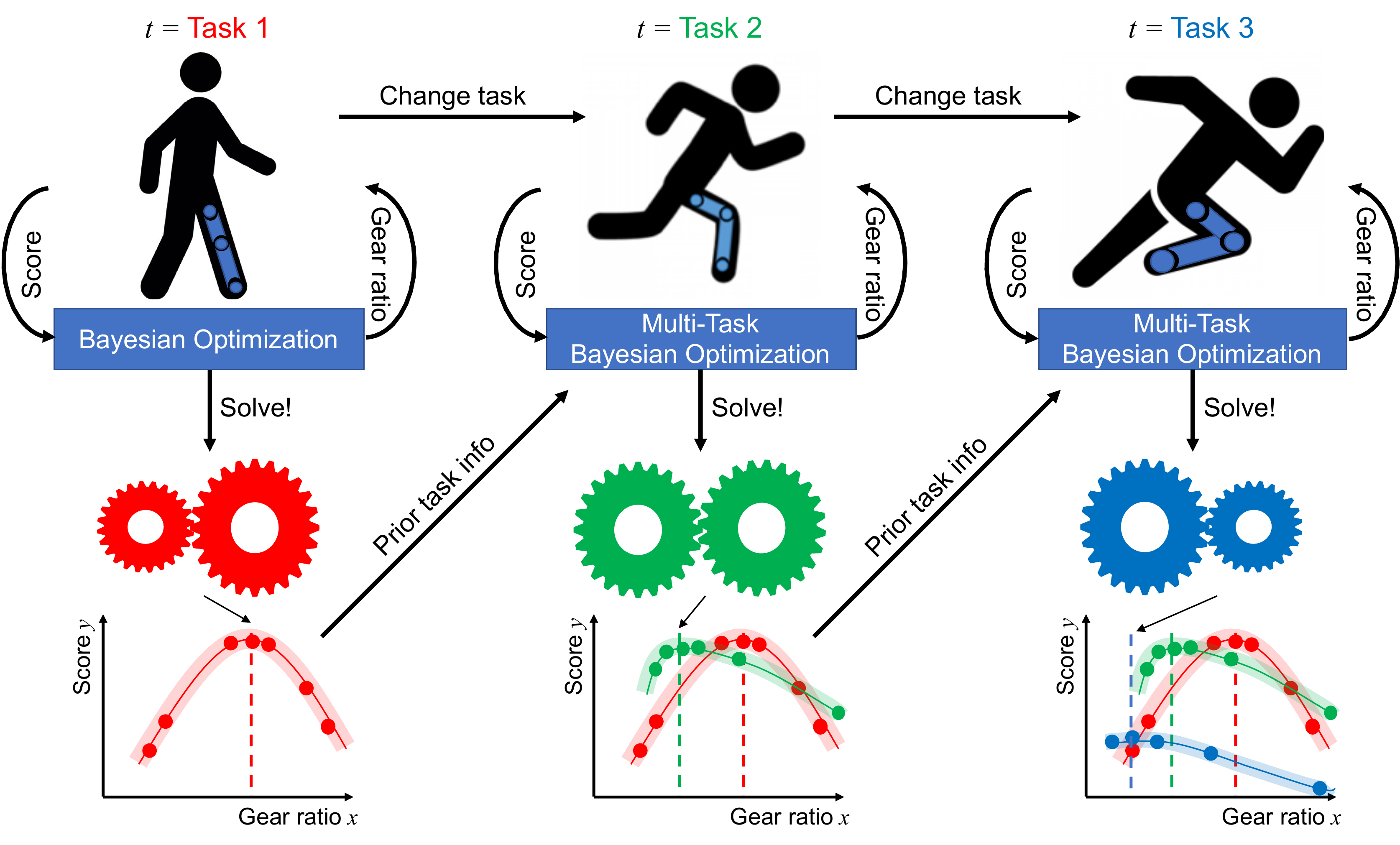}
    \caption{Rough sketch of sequential optimization for multiple tasks using multi-task Bayesian optimization:
    the data gained from previous tasks are reused for new tasks to predict the score functions according to similarities between the previous and new tasks.
    }
    \label{fig:BO_multi}
\end{figure}

The pseudo-code is shown in Algorithm~\ref{alg:bayes_opt}, and the rough sketch of sequential optimization for multiple tasks is shown in Fig.~\ref{fig:BO_multi}.
Here, since our framework considers multiple tasks (i.e., different environments and motions), how to infer the score function $\hat f(x; t)$ for each task is the critical issue to enhance new tasks' sample-efficiency by transferring the previous optimization results gained from different tasks.
Specifically, this issue is resolved by a multi-task Bayesian optimization~\cite{swersky2013multi} with a multi-task Gaussian process~\cite{bonilla2008multi}, details of which are given from the next section.

\begin{algorithm}[tb]
    \caption{Bayesian optimization with multiple tasks}
    \label{alg:bayes_opt}
    \begin{algorithmic}[1]
        \State{Black-box function $f$}
        \State{Dataset $\boldsymbol{D} \gets$ if available: $\{ \boldsymbol{X}, \boldsymbol{y}, \boldsymbol{t} \}$}
        \State{Prior $\gets$ if available: Prior of the score function model $\hat{f}$}
        \State{Number of iteration $i = 1$}
        \While{True}
        \State{// Learn $\hat{f}$}
        \State{Train $k^x$ and $K^f$ of $\hat{f}$ in eq.~\eqref{eq:bo_multi_cov} \\\qquad by maximizing the likelihood of $\boldsymbol{D}$ with the prior}
        \State{Compute Gram matrix $K$ from $k^x$ and $K^f$}
        \State{Set mean $\mu$ of $\hat{f}$ by eq.~\eqref{eq:BO_multi_mu}}
        \State{Set variance $\sigma^2$ of $\hat{f}$ by eq.~\eqref{eq:BO_multi_scale}}
        \State{// Get a new candidate of gear ratio $x^\prime$}
        \State{Set acquisition function $\alpha$ based on $\mu$ and $\sigma$ as eq.~\eqref{eq:ucb}}
        \State{Give the next task $t^\prime$}
        \State{Get the new candidate as $x^\prime = \arg \max \alpha(x; t^\prime)$}
        \State{// Evaluate $x^\prime$ and Update $\boldsymbol{D}$}
        \State{Get $y^\prime = f(x^\prime; t^\prime) + \epsilon$}
        \State{Add $\boldsymbol{D} \gets \{\boldsymbol{X} \gets x_i = x^\prime, \boldsymbol{y} \gets y_i = y^\prime, \boldsymbol{t} \gets t_i = t^\prime\}$}
        \State{// Check convergence}
        \If{$\sigma(\arg \max \mu(x; t^\prime); t^\prime)^2 < 5 \times 10^{-4}$ or \\\qquad the last three in $\boldsymbol{X}$ are the same}
        \State{\textbf{brak}}
        \EndIf
        \State{Iterate $i \gets i + 1$}
        \EndWhile
    \end{algorithmic}
\end{algorithm}

\subsection{Sequential multi-task gear ratio optimization}
\label{subsec:BO_multi}

Here, we show how to allow previous optimization results for different conditions about environments and motions to accelerate the new optimization process.
Our approach builds on the multi-task Gaussian process~\cite{bonilla2008multi}, where the correlations between multiple tasks are learned from data as their similarities.
Once we obtain the correlations, the knowledge of previous optimizations can be transferred to the new task optimization (details are in~\cite{swersky2013multi}).

Let's assume that the following data are collected from $M$ multiple tasks:
\begin{align}
    \boldsymbol X &= [x_1, x_2, \ldots, x_i, \ldots, x_T]
    \label{eq:dataset_x}\\
    \boldsymbol y &= [y_1, y_2, \ldots, y_i, \ldots, y_T]
    \label{eq:dataset_y}\\
    \boldsymbol t &= [t_1, t_2, \ldots, t_i, \ldots, t_T]
    \label{eq:dataset_t}
\end{align}
where $T$ denotes the number of data and the subscript $i = \{1, \ldots, T\}$ represents $i$-th data.
That is, in our case, $i$ specifies number of human-in-the-loop experiment for energy harvesting with $x_i$ gear ratio and $t_i$ task (e.g. walking on flat), and then, the experimental result is returned as $y_i$.

Assuming that $y_i = f(x_i; t_i) + \epsilon$ where $\epsilon \sim \mathcal{N} (0, \sigma^2_{t_i})$ with $\sigma_{t_i}$ the task-specific scale of noise, we place a Gaussian process prior over the $M$ latent functions $\{f(\cdot; m)\}_{m=1}^M$ so that we directly induce correlations between tasks as follows:
\begin{align}
    \mathrm{Cov}(f(x_i; t_i), f(x_j; t_j)) &= k(x_i, x_j, t_i, t_j)
    \nonumber\\
    &= K_{t_i t_j}^f k^x(x_i, x_j)
    \label{eq:bo_multi_cov}
\end{align}
where $j$ denotes another iteration number for $i$.
$k^x$ is a covariance function over inputs.
$K^f \in \mathrm{R}^{M \times M}$ is a positive semi-definite matrix that specifies the inter-task correlations, hence, its component $K_{t_i t_j}^f = K_{t_j t_i}^f$ specifies the similarities between $t_i$ and $t_j$.
These two, $k^x$ and $K^f$, can be learned by maximizing the marginal likelihood $p(\boldsymbol y \mid \boldsymbol X) = \mathcal{N} (0, K)$, where $K$ denotes the Gram matrix with $K_{t_i t_j}^f k^x(x_i, x_j)$ as $ij$-component.
That is, the inter-task correlation matrix $K^f$ can directly be optimized as part of GP optimization, without any other techniques for correlation analysis like canonical correlation analysis~\cite{hardoon2004canonical}.
In addition, we note that GP itself is not suitable for large scale data (also discussed in the section~\ref{subsec:discuss_efficiency}), hence the problem scale solved by our framework does not require the methods for finding sparse correlations in such large data~\cite{hardoon2011sparse}.

Given the new test input $x^\prime$ with its task label $t^\prime$, the predictive distribution of the corresponding output $y^\prime = \hat f(x^\prime; t^\prime)$ is given by $p(y^\prime \mid x^\prime; t^\prime, \boldsymbol X, \boldsymbol y, \boldsymbol t, k^x, K^f) = \mathcal{N}(\mu, \sigma^2)$ with $\mu$ location and $\sigma$ scale parameters defined as follows:
\begin{align}
    \mu(x^\prime; t^\prime) &= \boldsymbol k(x^\prime, t^\prime)^\top K^{-1} \boldsymbol y
    \label{eq:BO_multi_mu}\\
    \sigma(x^\prime; t^\prime)^2 &= k(x^\prime, x^\prime, t^\prime, t^\prime) - \boldsymbol k(x^\prime, t^\prime)^\top K^{-1} \boldsymbol k(x^\prime, t^\prime)
    \label{eq:BO_multi_scale}
\end{align}
where $\boldsymbol k(x^\prime, t^\prime)$ is the vector of covariances, namely $[k(x^\prime, x_1, t^\prime, t_1), k(x^\prime, x_2, t^\prime, t_2), \ldots, k(x^\prime, x_T, t^\prime, t_T)]^\top$.
Note that the derivation of these two equations has already been proved in~\cite{bonilla2008multi,swersky2013multi}.

Using the above parameters $\mu$ and $\sigma$, we can employ a popular acquisition function, so-called Upper Confidence Bound (UCB)~\cite{srinivas2010gaussian}:
\begin{align}
    \alpha(x; t) = \mu(x; t) + \kappa \sigma(x; t)
    \label{eq:ucb}
\end{align}
where $\kappa$ denotes the hyperparameter that controls the tradeoff between exploration and exploitation.

\begin{figure}[tb]
    \centering
    \begin{minipage}{0.49\linewidth}
        \centering
        \includegraphics[keepaspectratio=true,width=\linewidth]{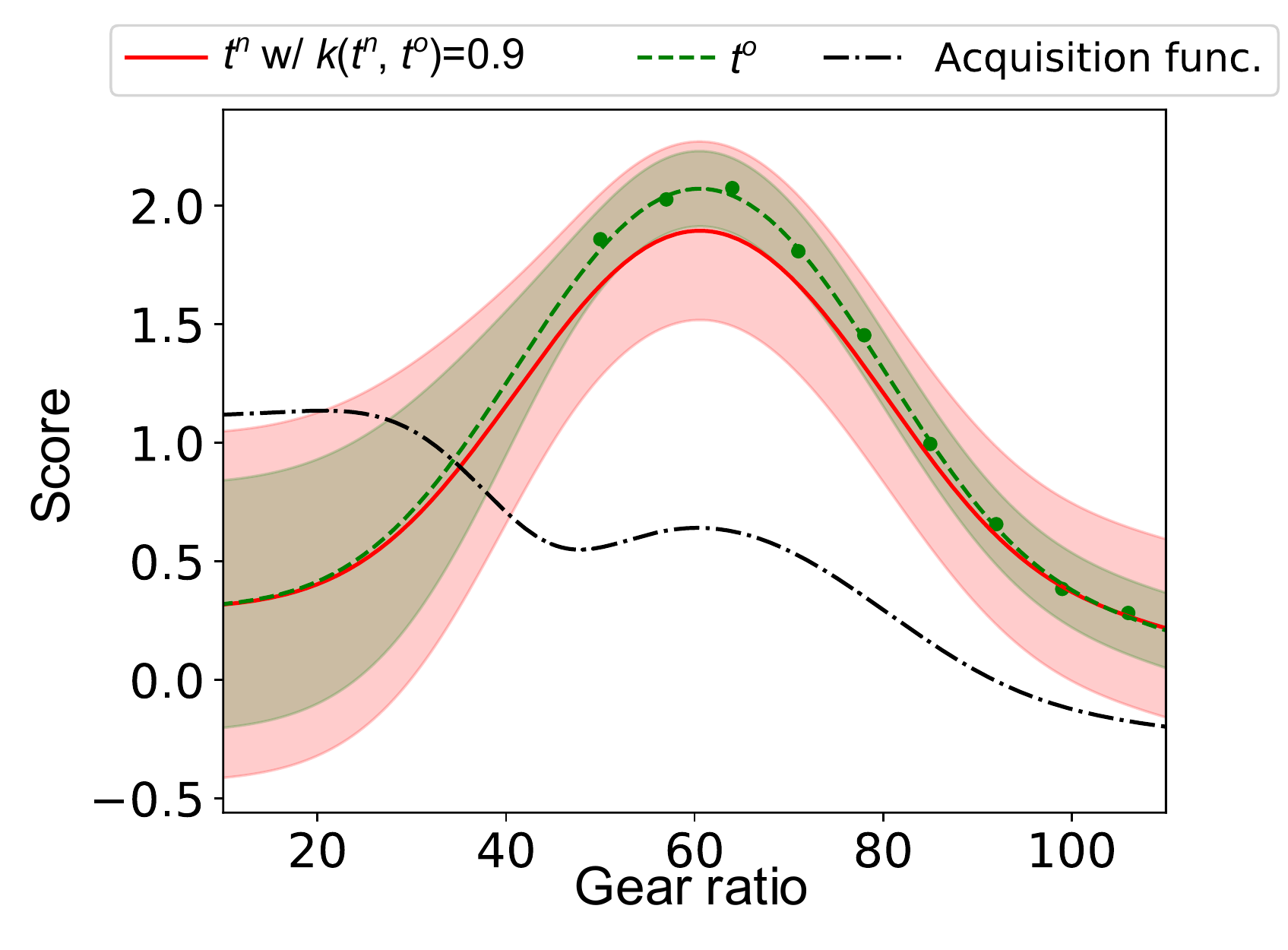}
        {(a) With strong correlation}
    \end{minipage}
    \begin{minipage}{0.49\linewidth}
        \centering
        \includegraphics[keepaspectratio=true,width=\linewidth]{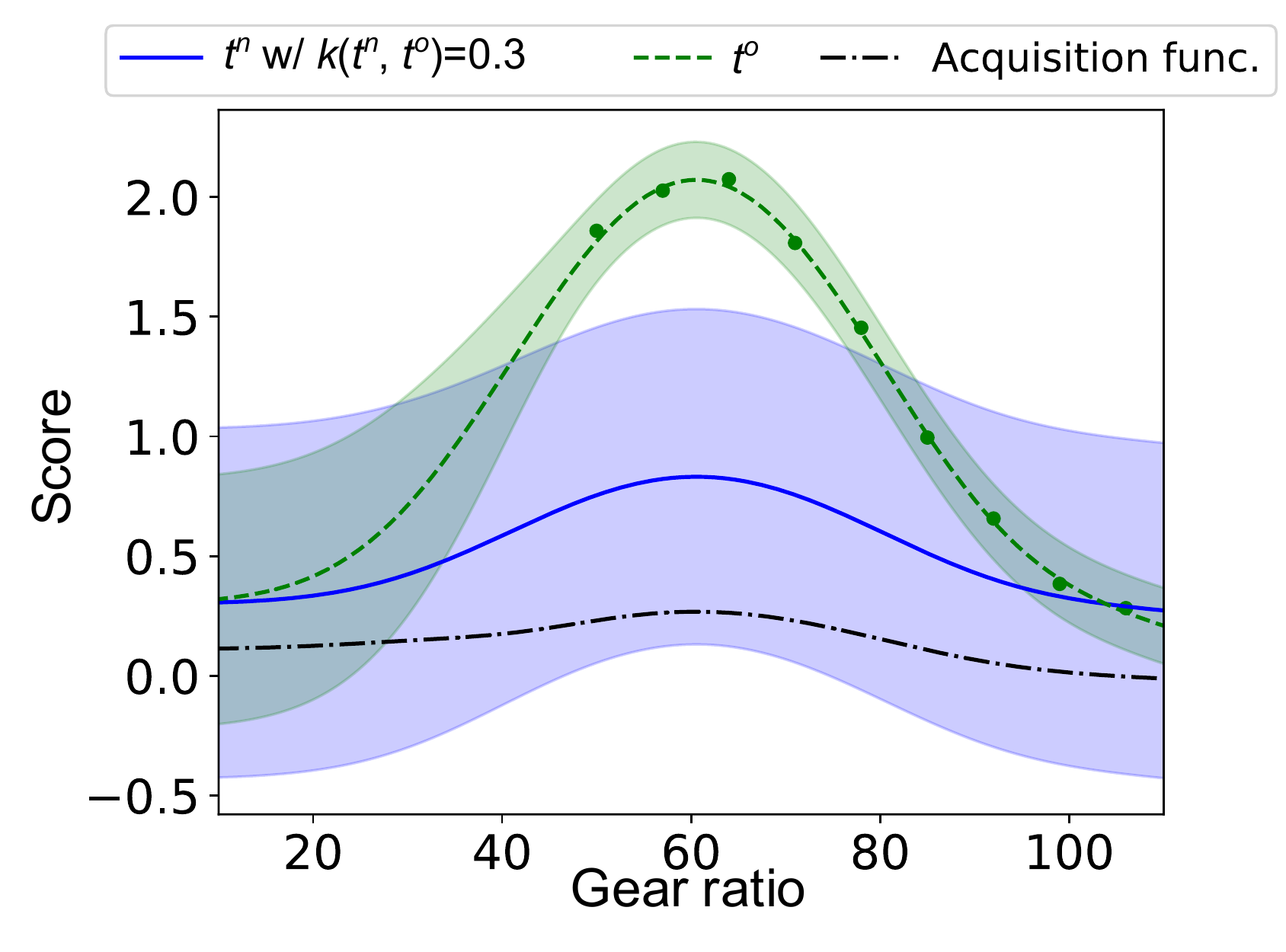}
        {(b) With weak correlation}
    \end{minipage}
    \caption{Illustration of multi-task GPs and acquisition functions for two tasks $t^o$ and $t^n$:
nine data points of task $t_{1,\ldots,9} = t^o$ are observed;
(a) when the new test-task label $t_{10} = t^n$ is set the correlation $k$ as $k(t^o,t^n) = 0.9$, the predictive distributions for the task $t^n$ exploit the data of tasks $t^o$;
(b) on the other hand, when $k(t^o,t^n) = 0.3$, it does not exploit the data much;
each solid line represents the mean, the shaded area represents the variance, and the black dashed line denotes the acquisition function.
    }
    \label{fig:multi_gp}
\end{figure}

Here, an illustrative example is shown in Fig.~\ref{fig:multi_gp}.
If there are strong correlations between the previous and new tasks ($t^o$ and $t^n$ respectively), this transfer makes the optimization speed-up.
Indeed, as shown in Fig.~\ref{fig:multi_gp}(a), the predicted score functions for the previous and new tasks almost overlap, and the acquisition function restarts the exploration for low gear ratio, which is not explored yet in the previous task.
On the other hand, if the new task has a weak correlation with the previous one (see Fig.~\ref{fig:multi_gp}(b)), the predicted score function for the new task mostly ignores the one for the previous task, and the acquisition function almost starts the exploration from scratch.
Note that, even in this case, this approach does not deteriorate the performance and turns out to be just a single task optimization problem.

\section{Experimental Setup}
\label{sec:setup}

\begin{figure}[tb]
    \centering
    \includegraphics[keepaspectratio=true,width=0.8\linewidth]{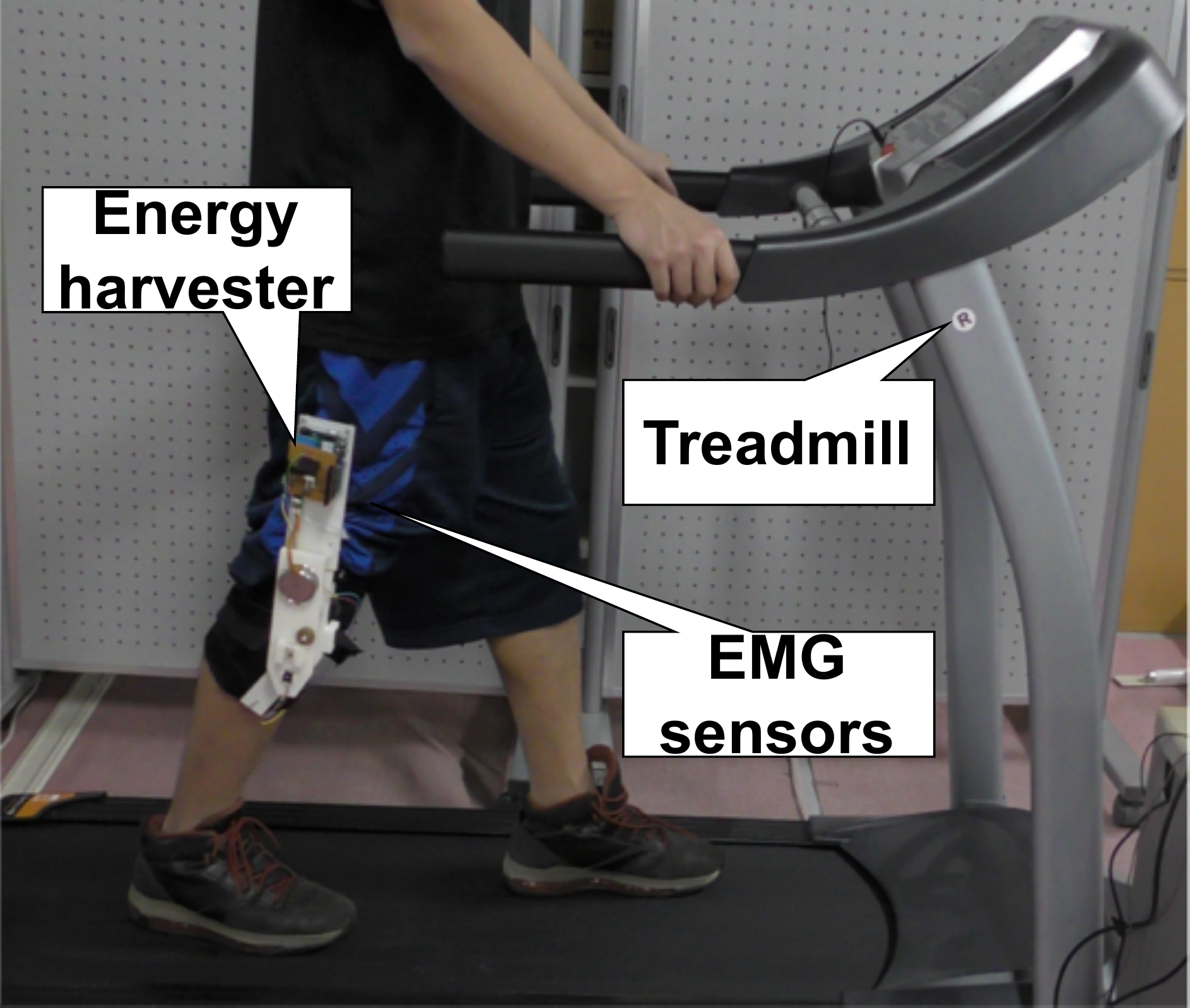}
    \caption{Experimental setup:
    two participants equipped with the the energy harvester walk (or jog) on a treadmill with multiple slope angles and multiple speeds;
    the burden of the user is estimated by the sensation of EMG sensors.
    }
    \label{fig:exp_env}
\end{figure}

We verify the effectiveness of the proposed framework in real-world experiments where two participants equipped with the energy harvester walk (or jog) on a treadmill with multiple slope angles and multiple speeds (see Fig.~\ref{fig:exp_env}).
Here, the experimental purpose only focuses on the sample-efficiency of our framework to find the optimal gear ratio, which is obtained through brute force search in advance.
All the experiments were conducted under the approval of the Ethics Committee of Nara Institute of Science and Technology.

\subsection{Harvesting timing focusing on brake motion}

To efficiently harvest energy against the burden of the user, the generator and the circuit are connected only during a brake motion (braking of swing leg in the gait, specifically), as mentioned before.
This duration is heuristically defined by a potentiometer to measure the knee angle: the duration when the angle is under 15$^\circ$ and the angular velocity is negative.
Note that, to emphasize the effects of CVT, the duty cycle $D$ introduced in eqs.~\eqref{eq:watt} and~\eqref{eq:torque} is limited from $[0, 1]$ to binary $\{0, 1\}$, which means whether the circuit is connected or not.

\subsection{Measurement systems of evaluation score}
\label{subsec:score}

\subsubsection{Generated power}

Instead of the generated power, $W$, the counter-electromotive force rectified by a three-phase full-wave rectifier circuit, $V$, is measured through an A/D converter of Arduino Uno.
It can theoretically be converted into $W$ if the internal resistance $R$ is known: $W = V^2/R$.
Note that $W$ is estimated after applying a low-pass filter to $V$ for noise removal.
This low-pass filter is implemented in Arduino Uno as a moving average based filter.
Although this estimation of $W$ is not accurate due to the different $R$ when the actual battery is connected, its tendency is correct.
In the experiments, this approximation is reasonable since we only focus on the sample-efficient optimization of the gear ratio without considering the true power generation.

\subsubsection{Burden of the user}

As a criterion of the user's burden, we employ electromyography (EMG).
The target motion of the energy harvesting in the experiments is the brake motion at the time of swing-leg retraction in the gait.
Hence, the EMG sensors (Oisaka Development Ltd.: P-EMG plus) are attached on vastus lateralis muscle and vastus medialis muscle related to the target motion (i.e., knee extensors at the swing-leg retraction).
Full-wave rectification smoothing enables us to extract the amount of the muscle activity since raw values of EMG include negative values.
The EMG signals would become large if the anti-torque during harvesting hinders the motion or the brake.

\subsubsection{Score function}

In our harvester system, the input $x$ for the optimization is equivalent to the gear ratio $G$ in CVT.
In that case, the score $f(G; t)$ to be maximized is heuristically defined from the generated power and the EMG as follows:
\begin{align}
	f(G;t ) = W(G; t) - \gamma \mathrm{EMG}(G; t)
    \label{eq:score}
\end{align}
where $\gamma$ is the hyperparameter to combine two different unit terms, and to satisfy the requirement in which the effects to the user's motion should be minimized.
It is set as $\gamma = 5$ in all the experiments empirically.
Although this score can also be measured as an instantaneous value, the average value in each trial is taken as the actual score to suppress the influence of motion noise.

It is relatively easy to measure both $W$ and $\mathrm{EMG}$ under a certain value of $G$ through experiments.
However, its exact functions are difficult to derive mathematically since the angular velocity of the knee $\dot{\theta}_k$ in eqs.~\eqref{eq:watt} and~\eqref{eq:torque} is unclearly affected by $G$ and the target task $t$.
That is why the black-box optimization approach is required to efficiently find the optimal gear ratio from a few number of trials.

\subsection{Experimental conditions for data collection}

\begin{table}[tb]
    \centering
    \small
    \caption{Specification of participants}
    \begin{tabular}{ c r r r r }
        \hline
        Participant & Height [m] & Weight [kg] &  Inseam [m]& Age
        \\ \hline
        1 & 1.74 & 75.2 & 0.85 & 24
        \\
        2 & 1.70 & 60.0 & 0.80 & 23
        \\ \hline
    \end{tabular}
    \label{tab:params_participant}
\end{table}

\begin{figure}[tb]
    \centering
    \begin{minipage}{0.49\linewidth}
        \centering
        \includegraphics[keepaspectratio=true,width=\linewidth]{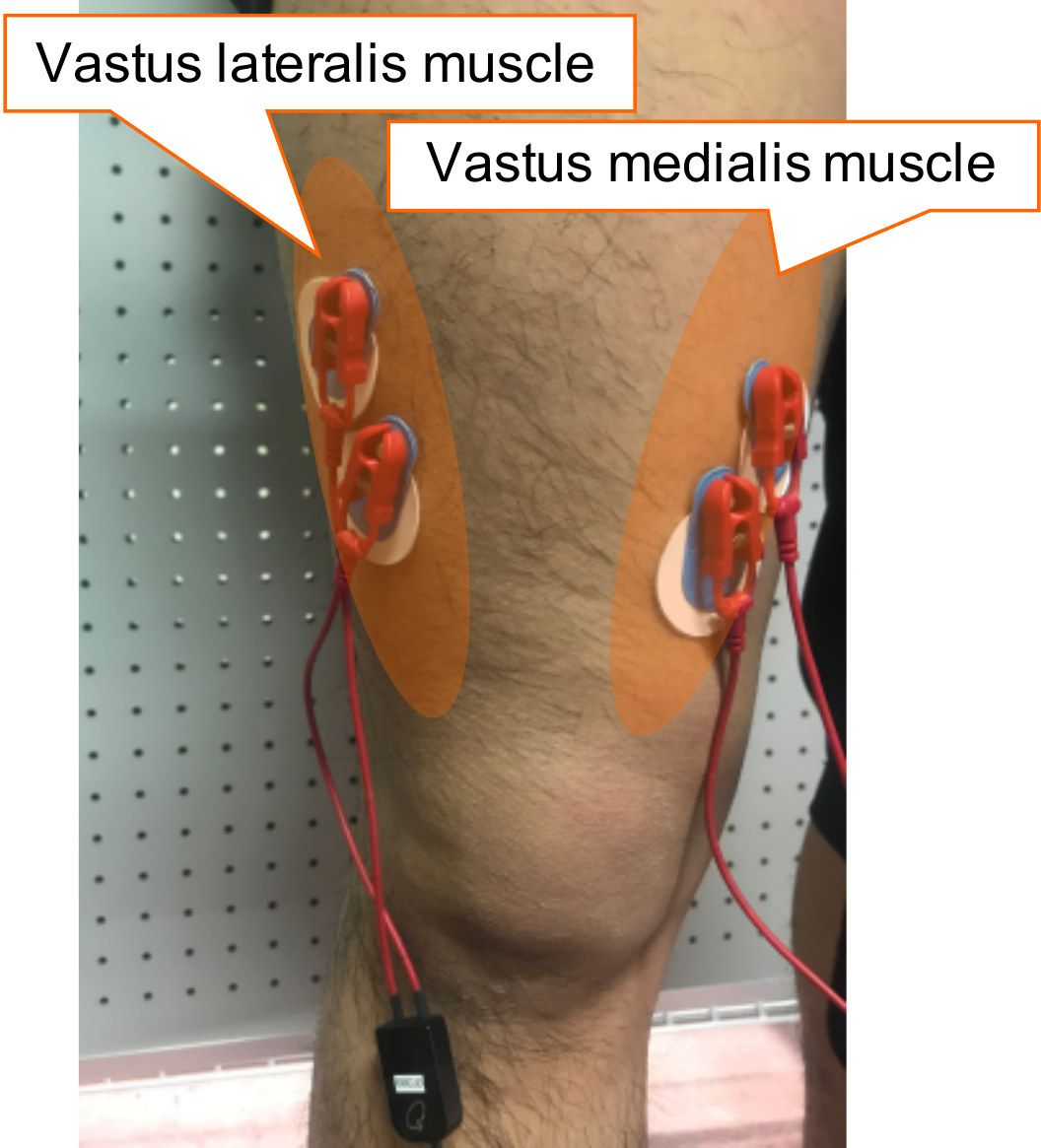}
        {(a) EMG sensors}
    \end{minipage}
    \begin{minipage}{0.49\linewidth}
        \centering
        \includegraphics[keepaspectratio=true,width=\linewidth]{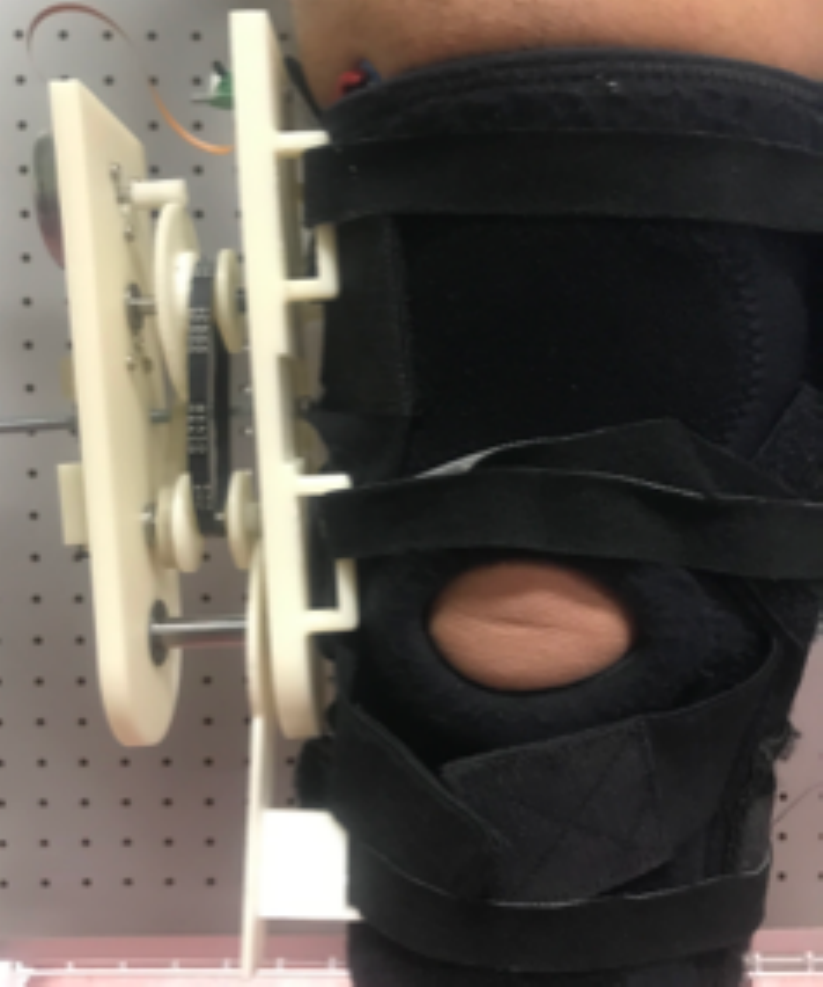}
        {(b) Energy harvester}
    \end{minipage}
    \caption{Attachment of the devices for the experiments:
    (a) the EMG sensors are attached on vastus lateralis muscle and vastus medialis muscle in the right leg;
    (b) the energy harvester is attached on the EMG sensors.
    }
    \label{fig:exp_attach}
\end{figure}

To reveal the relationship between the score and the gear ratio, we first collect the scores w.r.t all the candidates of gear ratio (50 equal discretizations of $[16, 144]$ in this paper) under multiple conditions (i.e., the tasks).
Two participants summarized in Table~\ref{tab:params_participant} walk (or jog) on the treadmill with the gait speed from $\{1, 1.5, 2\}$ m/s and the slope angle from $\{0, 5, 10\}$ deg.
That is, $50 \times 2 \times 3 \times 3 = 900$ data are sampled in total.
The collected data is used for illustrating ground truths for the respective tasks.

Let us describe the protocol of one trial.
At first, a participant experiences the target task on the treadmill for 30 seconds before evaluation to get used to gait under the target task.
After that, the generated power and the EMG are measured only from 10 seconds to 20 seconds after starting the gait to ignore data measured from non-steady gait.
The score under the target task is evaluated as the average of the measured values for that 10 seconds.
An interval of 70 seconds is inserted before the next trial is started.
During this time, a participant relaxes on a seat.
In total, one trial needs 120 seconds (2 minutes).

The above protocol is repeated 150 times (50 gear ratios under three types of slope angles) in one day.
Namely, we take three days for the experiments per participant.
The attached positions of the EMG sensors are marked on the participant's skin directly to keep the measurement positions of the EMG fixed, as shown in Fig.~\ref{fig:exp_attach}.
When the EMG sensors are re-attached the next day, their positions are fine-tuned by comparing the output values from a calibration behavior that day and the previous day.

\subsection{Exploration of the optimal gear ratio}

Using the obtained data, we investigate the effectiveness and sample-efficiency of our framework for exploring the optimal gear ratio.
In particular, we consider two scenarios for optimizing gear ratios as:
\begin{itemize}
    \item \textit{multiple slope angles} ($\{0, 5, 10\}$ deg) under the respective speeds ($\{1, 1.5, 2\}$ m/s)
    \item \textit{multiple gait speeds} ($\{1, 1.5, 2\}$ m/s) under the respective slope angles ($\{0, 5, 10\}$ deg)
\end{itemize}
Thus, acceleration effects of exploration using the correlations among different slope angles or different speeds are investigated respectively.

Exploration of the optimal gear ratio is terminated when either of the following two conditions is satisfied:
\begin{itemize}
    \item Predictive variance on the maximum mean is under $5 \times 10^{-4}$.
    \item The same gear ratio is selected three times sequentially.
\end{itemize}
As mentioned above, the acquisition function to find the next candidate is given as UCB described in eq.~\eqref{eq:ucb}.
The hyperparameter $\kappa$ is set as $100$ in the experiments to emphasize exploration to avoid local optima rather than exploitation.

\begin{figure*}[tb]
    \centering
    \begin{minipage}{0.49\linewidth}
        \centering
        \includegraphics[keepaspectratio=true,width=\linewidth]{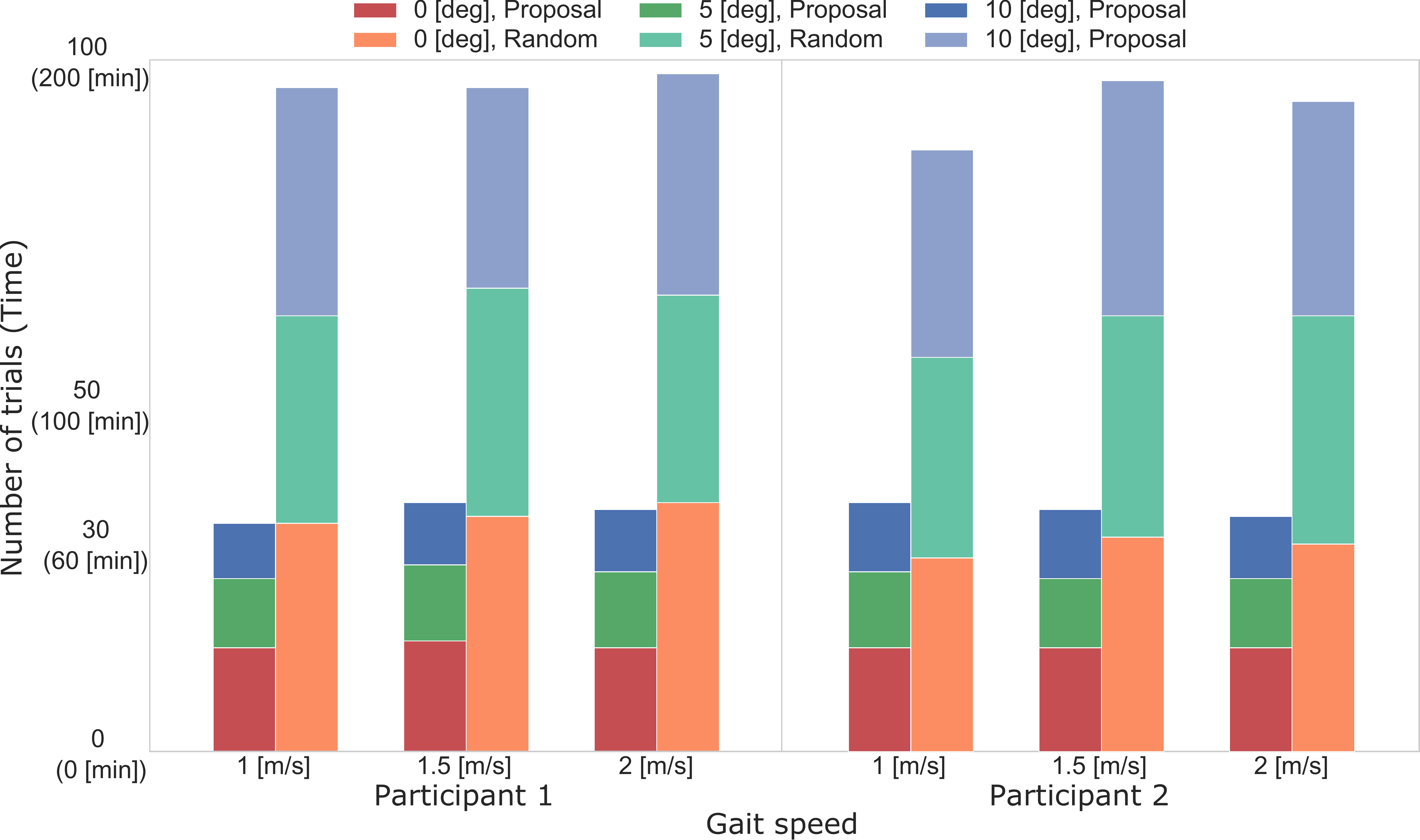}
        {(a) Multiple slope angles}
    \end{minipage}
    \begin{minipage}{0.49\linewidth}
        \centering
        \includegraphics[keepaspectratio=true,width=\linewidth]{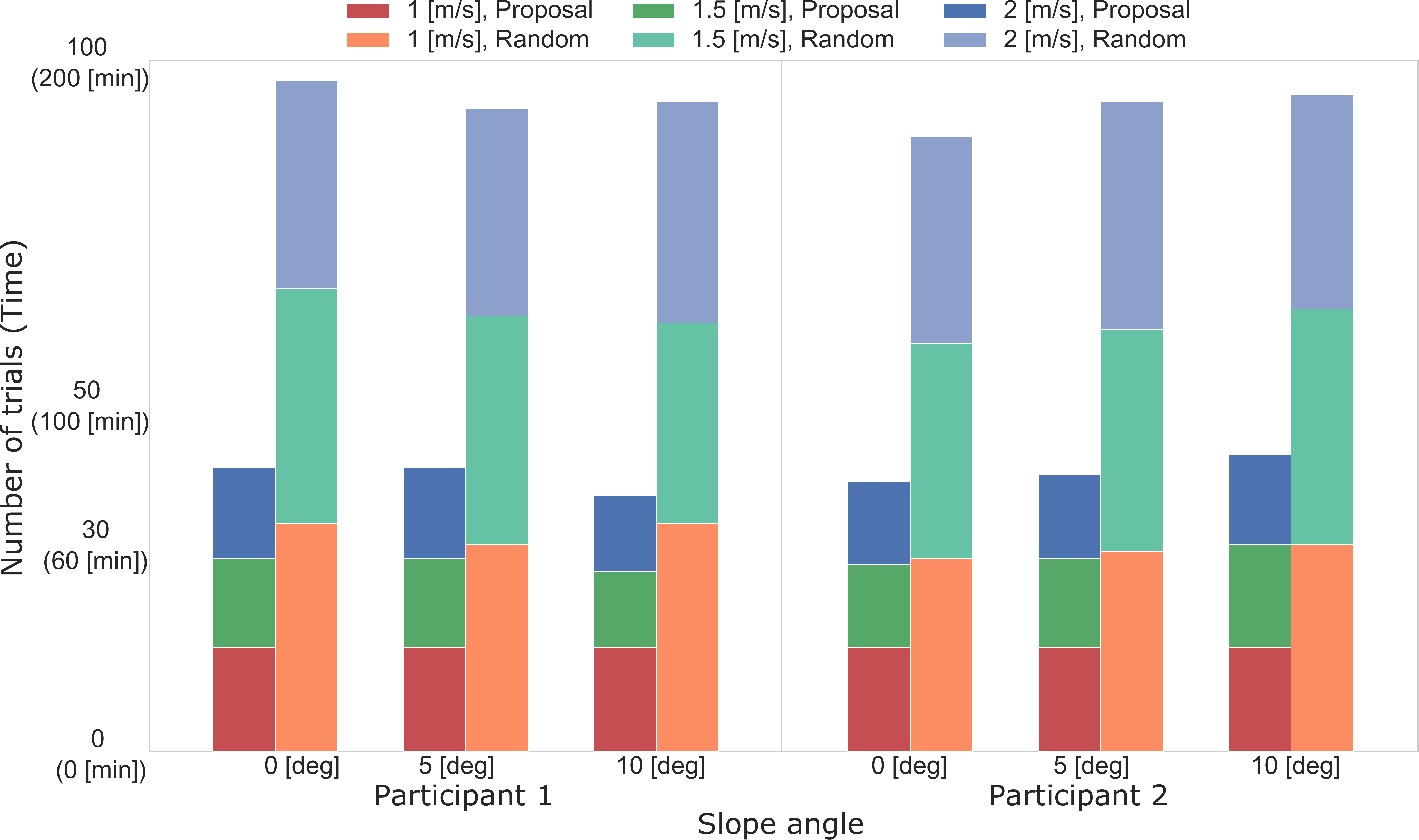}
        {(b) Multiple gait speeds}
    \end{minipage}
    \caption{Comparisons of the number of explorations:
    the proposed framework (a.k.a., Proposal) and a random search (a.k.a., Random) are compared;
    under all the scenarios, our proposed framework reduced the number of trials to search the optimal gear ratio to more than half of the random search, and the optimization for the three tasks in each scenario was completed in approximately one hour;
    in particular, the second and third tasks to be optimized were accomplished in a shorter time by utilizing similarities with the previous tasks.
    }
    \label{fig:exp_result_number}
\end{figure*}

\begin{figure*}[tb]
    \centering
    \includegraphics[keepaspectratio=true,width=\linewidth]{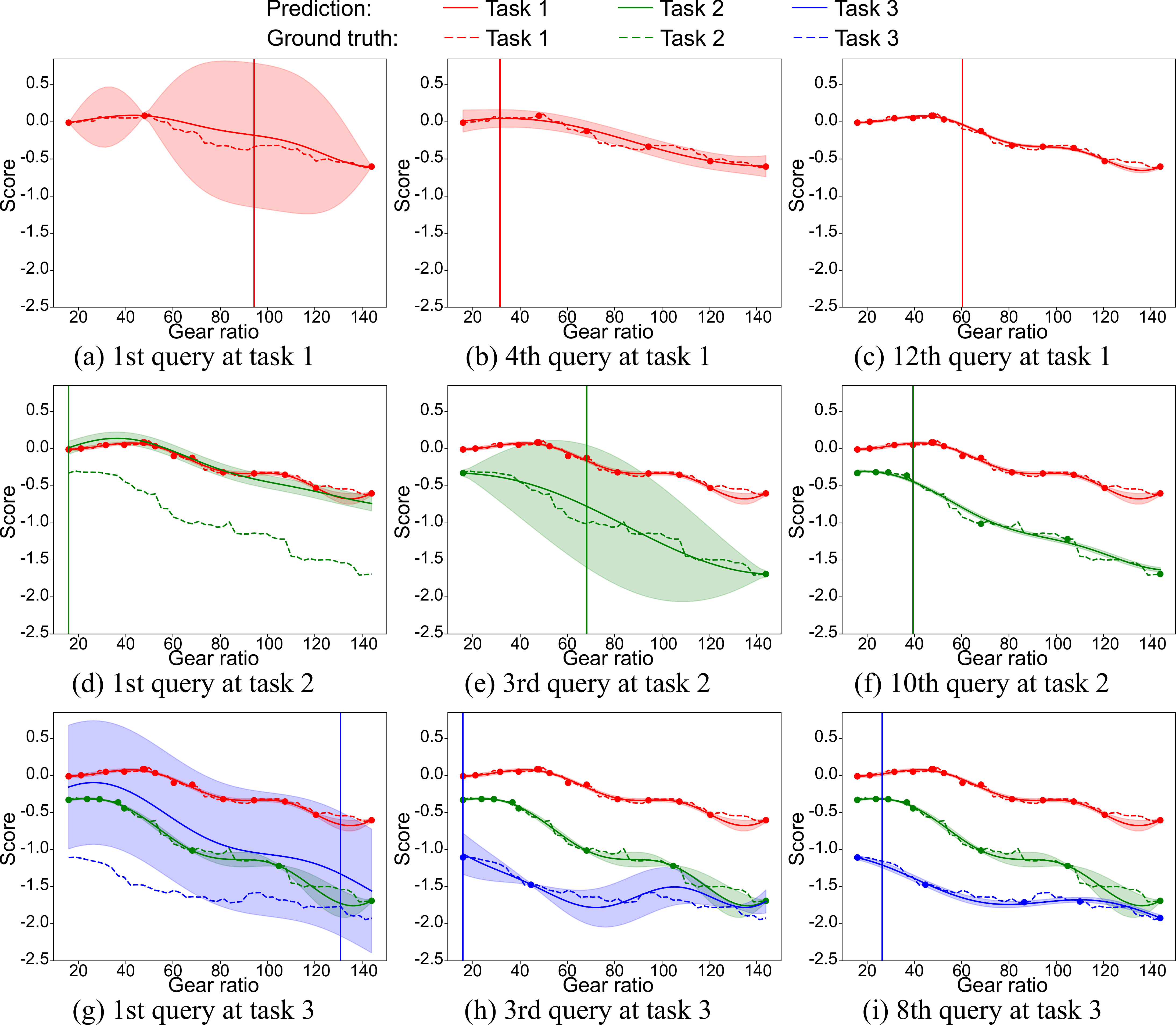}
    \caption{Exploration behaviors of our proposed framework:
    these behaviors were gained from the participant 1 with 1~m/s on 0 (red), 5 (green), and 10 (blue) deg slope angles, respectively;
    the dashed lines are ground truths of the corresponding tasks;
    the solid lines and the corresponding shadows are predicted means and variances of Gaussian processes, respectively.
    }
    \label{fig:exp_result_behavior}
\end{figure*}

\begin{table*}[tb]
    \centering
    \caption{Optimized gear ratios (multiple slope angles) with the ground truths shown in blankets:
    bolds mean the cases where the optimized gear ratios match the ground truths;
    the optimal gear ratios were found under 15 conditions out of 18 conditions;
    even in the failure cases, the near-optimal gear ratios were selected.
    }
    \begin{tabular}{|c|c|r|r|r|}
        \hline
        Participant & Slope & \multicolumn{3}{|c|}{Gait speed}
        \\ \cline{3-5}
        & & 1~m/s & 1.5~m/s &  2~m/s
        \\ \hline
        & 0~deg & \textbf{ 49.8 (49.8)} & \textbf{ 128.3 (128.3)}& \textbf{ 138.7 (138.7)}
        \\ \cline{2-5}
        1 & 5~deg & \textbf{ 26.3 (26.3)}&\textbf{ 109.9 (109.9)}&\textbf{ 117.8 (117.8)}
        \\ \cline{2-5}
        & 10~deg & \textbf{ 16.0 (16.0})& 73.3 (75.9)& \textbf{ 102.1 (102.1)}
        \\ \hline
        & 0~deg & \textbf{ 34.1 (34.1)}&\textbf{ 112.6 (112.6)}&\textbf{ 130.9 (130.9)}
        \\ \cline{2-5}
        2 & 5~deg & \textbf{ 23.6 (23.6)}&  107.3 (109.9)& \textbf{ 120.4 (120.4)}
        \\ \cline{2-5}
        & 10~deg & \textbf{ 18.4 (18.4)}& \textbf{ 65.5 (65.5)}&  112.6 (115.2)
        \\ \hline
    \end{tabular}
    \label{tab:exp_result_opt}
\end{table*}

\begin{figure}[tb]
    \centering
    \begin{minipage}[]{0.55\linewidth}
        \centering
        \includegraphics[keepaspectratio=true,width=\linewidth]{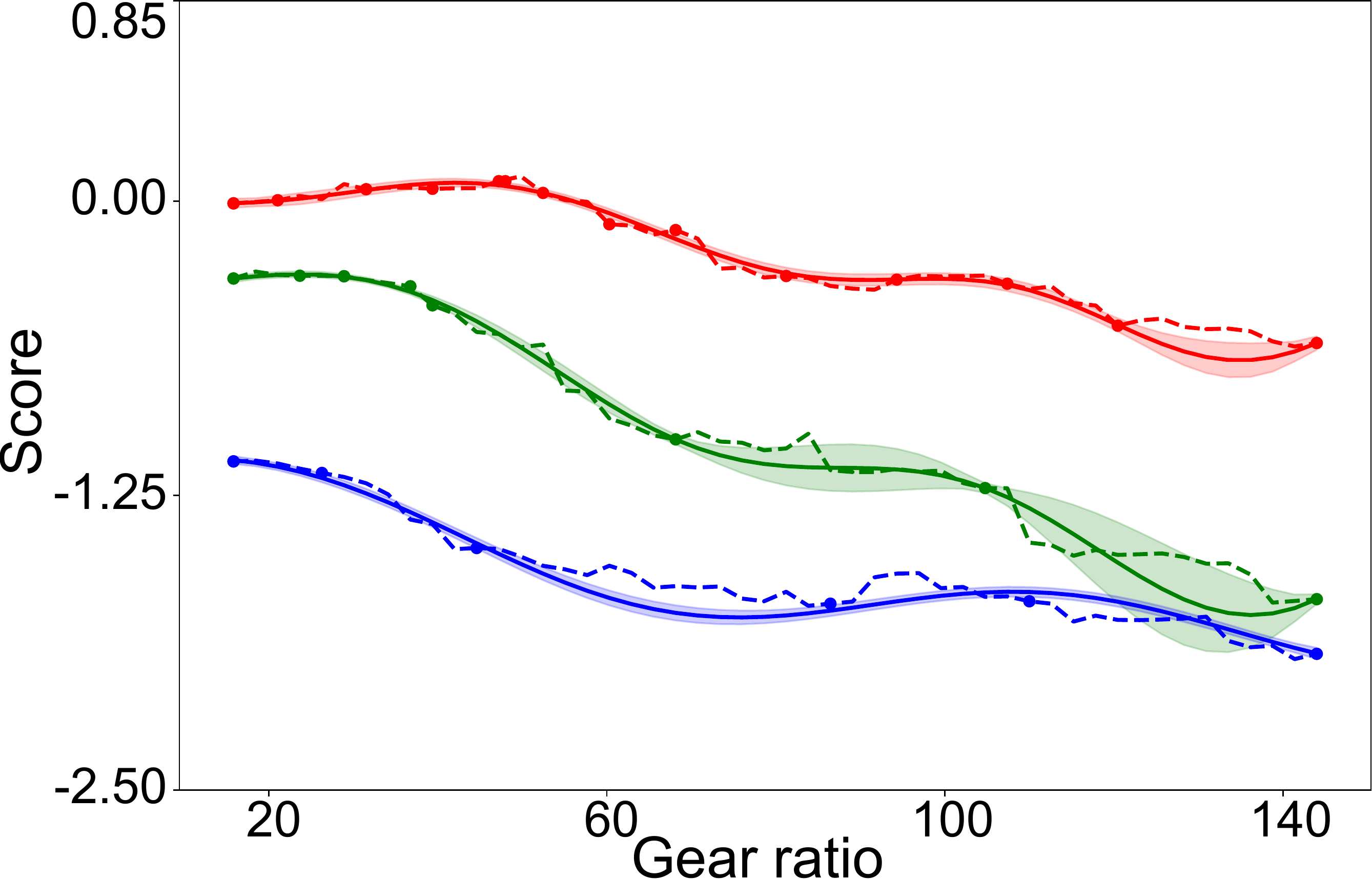}
        {(a) Participant 1 with 1~m/s}
    \end{minipage}
    \begin{minipage}[]{0.4\linewidth}
        \centering
        \includegraphics[keepaspectratio=true,width=\linewidth]{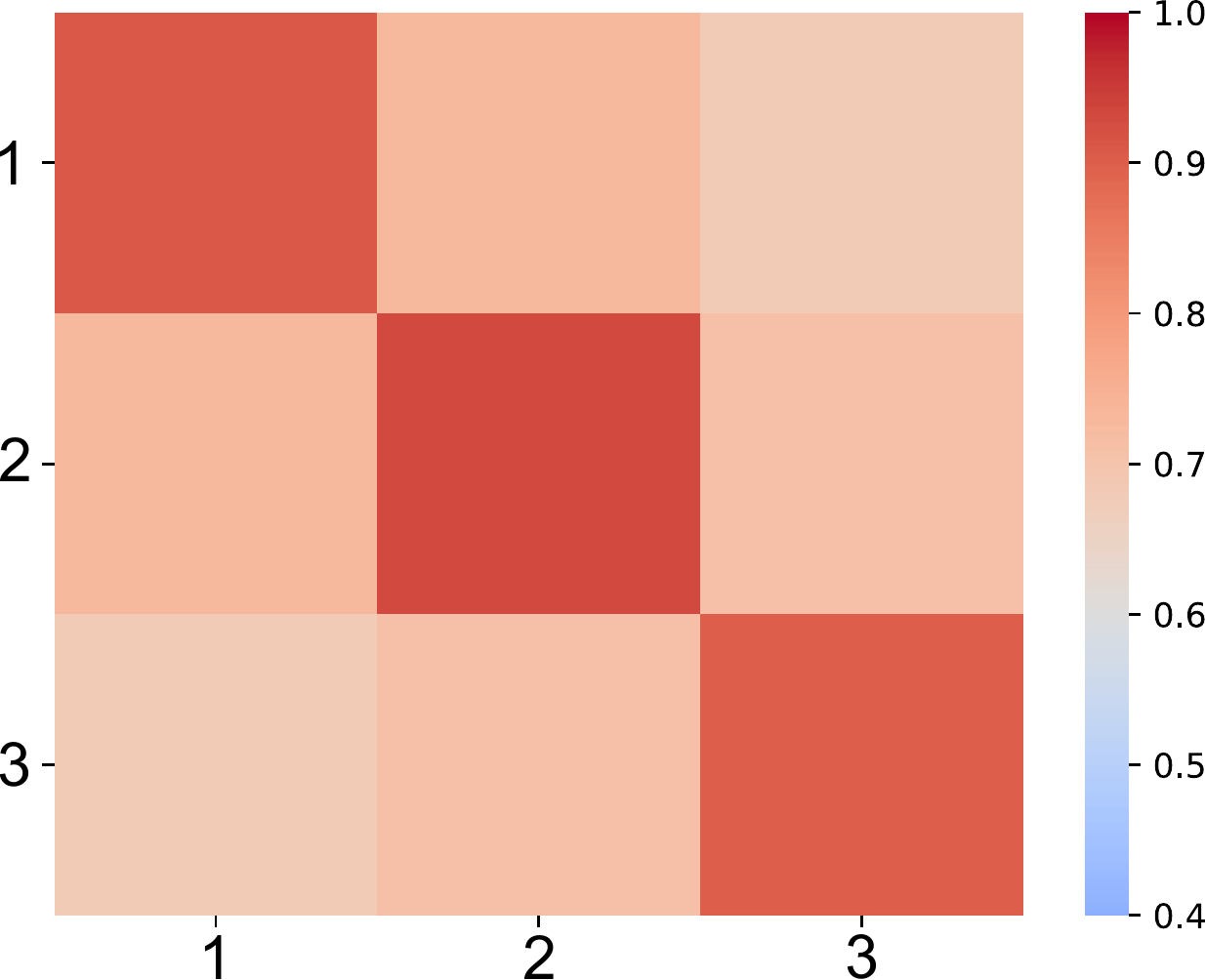}
        {(b) Correlation matrix $K^f$ of (a)}
    \end{minipage}
    \begin{minipage}[]{0.55\linewidth}
        \centering
        \includegraphics[keepaspectratio=true,width=\linewidth]{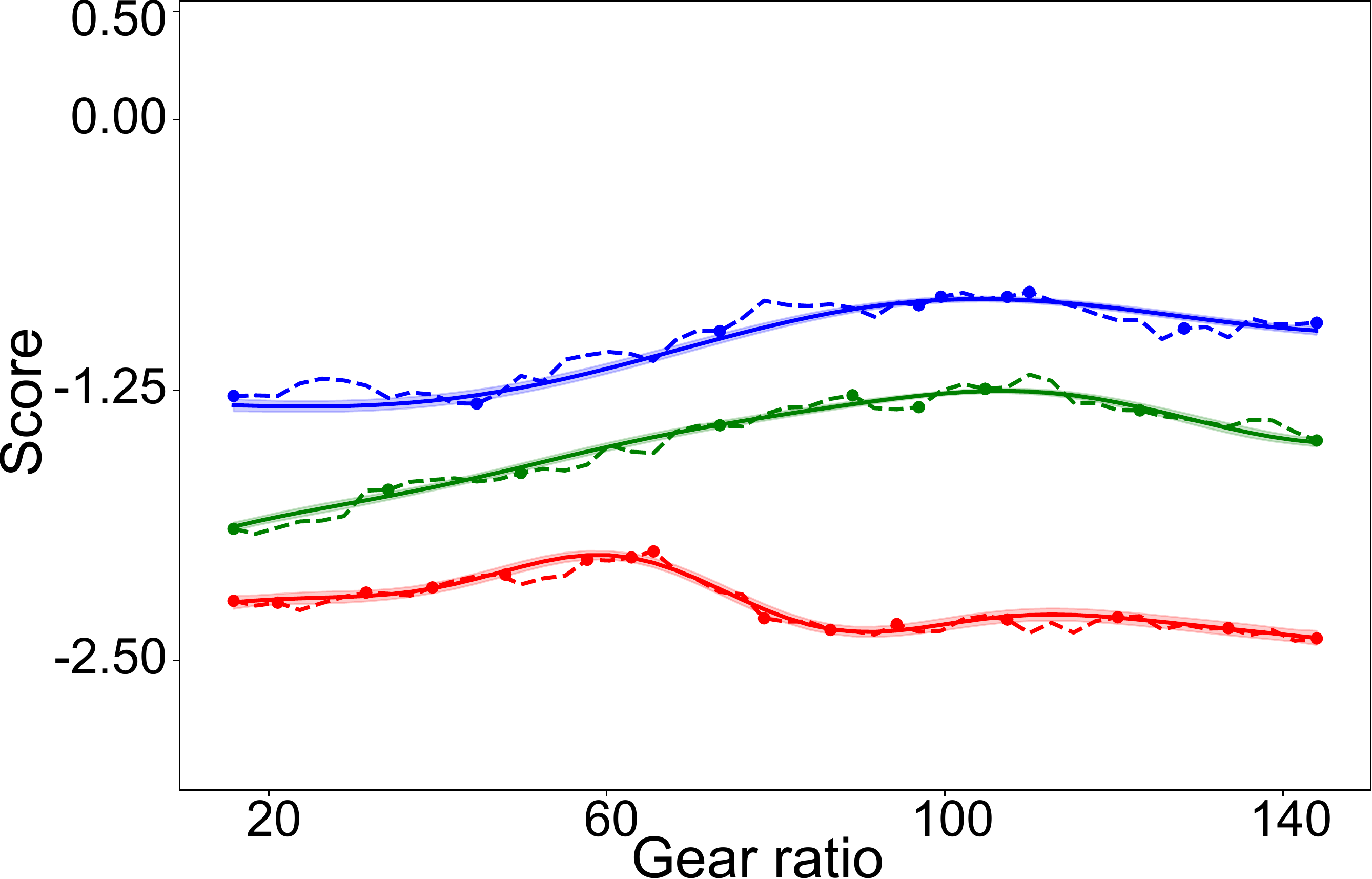}
        {(c) Participant 2 with 1.5~m/s}
    \end{minipage}
    \begin{minipage}[]{0.4\linewidth}
        \centering
        \includegraphics[keepaspectratio=true,width=\linewidth]{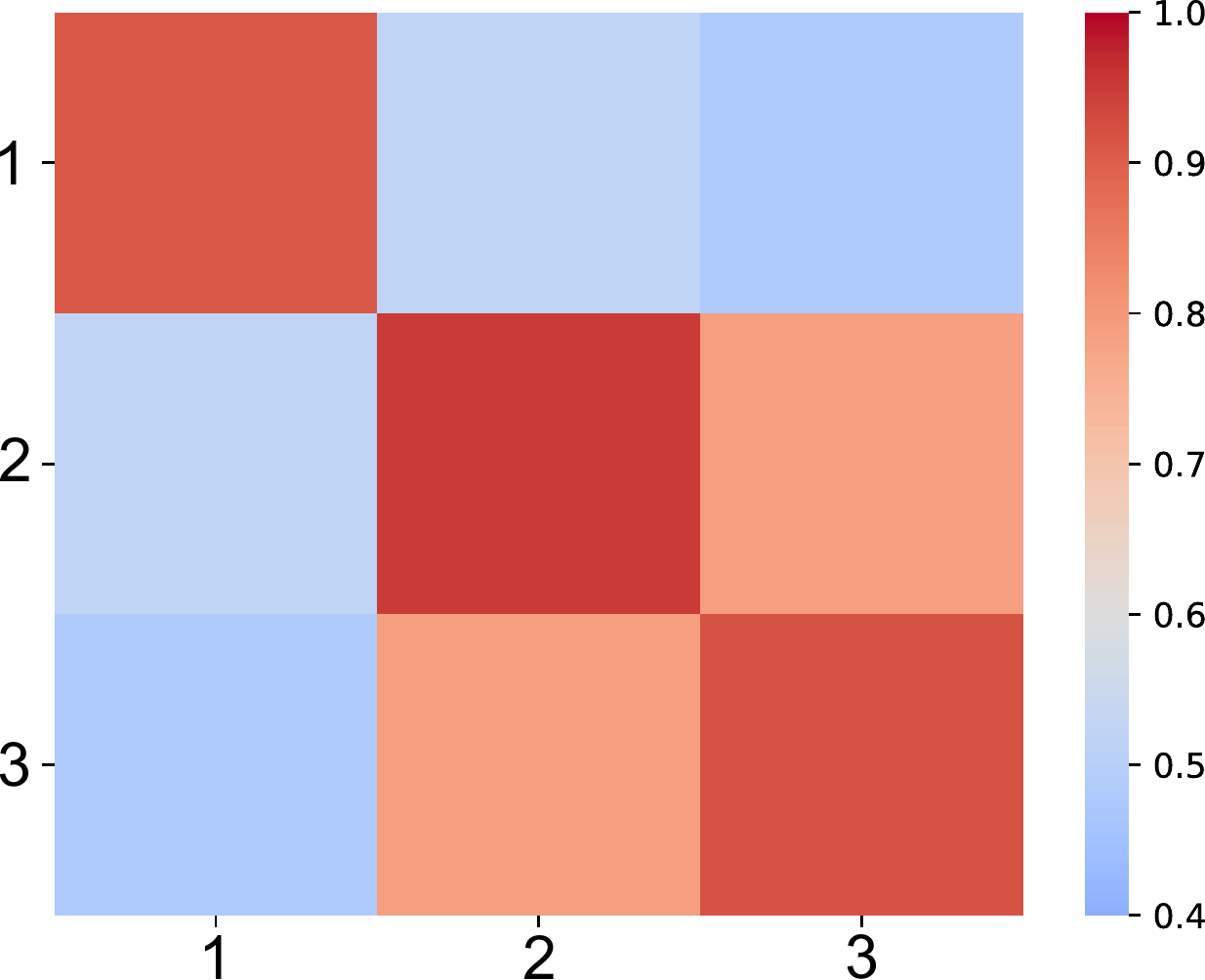}
        {(d) Correlation matrix $K^f$ of (c)}
    \end{minipage}
    \caption{Optimized task kernels:
    red, green, and blue lines are consistent with task 1, 2, and 3, respectively;
    (a)--(b)/(c)--(d) are the cases of high/low correlations between the tasks.
    }
    \label{fig:exp_result_correlation}
\end{figure}

\section{Experimental Results}
\label{sec:result}

\subsection{The number of explorations}

To verify our framework's sample-efficiency, the number of explorations (i.e., the time for optimization) is compared to a random search using the discrete uniform distribution as a baseline.
Specifically, our framework was applied to 12 split groups in total, as shown in Fig.~\ref{fig:exp_result_number}.
In all the cases, the explorations of our approach were terminated earlier than that of the baseline.
Our approach decreased the number of required trials as the number of tasks increased, although the amount of reduction was dependent on the groups (i.e., the correlations between tasks).
We found that the correlations between motions on different slope angles were higher than that between motions with different speeds commonly for both participants.
As a result, although the random search required around 60~minutes for the optimization per task, our framework took only around 20~minutes.

\subsection{Exploration behaviors}

The exploration behaviors in the scenario of multiple slope angles, where participant 1 walked with task 1) 1 m/s speed on 0 deg slope; task 2) 5 deg slope; and task 3) 10 deg slope, are depicted as one example (see Fig.~\ref{fig:exp_result_behavior}).
The exploration on task 1 was actually the same as standard Bayesian optimization. Namely, the predicted score function was entirely given by the sampled data only from task 1.
From task 2, the sampled data from the previous tasks were exploited, and therefore, the initial predictions shown in Figs.~\ref{fig:exp_result_behavior}(d) and~(g) had similar shapes to the previous predictions.
Thanks to the data sampled from the previous tasks, the exploration on task 3 was terminated with around half the number of queries (i.e., eight queries on task 3 compared to 12 (+ 3 for initialization) queries the task 1).
The predicted score functions finally gained were mostly consistent with the ground truths, even with fewer samples from task 2 and task 3.

\subsection{Quality of optimized gear ratios}

Table~\ref{tab:exp_result_opt} summarized all the optimization results with the ground truths in brackets for the scenario of multiple slope angles.
Our approach successfully found the optimal gear ratios under 15 conditions out of 18 conditions.
Even in unmatched cases, the results were mostly close to the optima.
These results suggest that our approach could find near-optimal gear ratios suitable for any tasks in a data-driven manner.

Furthermore, we can see the relationships between tasks qualitatively.
For instance, the increase of the slope angle or the gait speed always decreased or increased the optimal gear ratios, respectively.
Such relationships are reasonable.
That is, the slope angle affects only the user's burden, which should be reduced.
In addition, the gait speed sets the corresponding angular velocity of the knee joint, 2nd power of which is proportional to the generated power (see eq.~\eqref{eq:watt}). Hence, our approach prioritized the generated power rather than the user's burden.
Exploiting such relationships would facilitate finding the optimal gear ratios, in other words, that would make our approach further efficient. .

\subsection{Analyzing the optimized task kernels}

Here, we analyze the optimized task kernels (i.e., $K^f$ in eq.~\eqref{eq:bo_multi_cov}) and their effects on explorations.
Fig.~\ref{fig:exp_result_correlation} depicts the predicted score functions and the corresponding task kernels.
When high correlations between tasks were found as illustrated in Figs.~\ref{fig:exp_result_correlation}(a) and~(b), the explorations for the new tasks (the task 2 and 3) were much focused at around on the maximum since the landscape of the score function can be predicted from the data of the previous tasks even in the early trials.
Thus, the number of explorations was kept small.
Otherwise, as illustrated in Figs.~\ref{fig:exp_result_correlation}(c) and~(d), the explorations of the gear ratio for the new tasks were executed extensively across the entire domain.
Even in that case, since our proposed framework would revert to the standard Bayesian optimization, the deterioration of the sample-efficiency does not happen.

\section{Discussion}
\label{sec:discussion}

\subsection{Design of score function}

The score function to be optimized in eq.~\eqref{eq:score} aims to maximize the generative power and minimize the EMG on average.
However, it is hard to regard ours as \textit{the optimal score function} due to its ad-hoc design.
Here, we have to discuss the candidates of the score function.
Finally, the optimal one should be selected from them according to meta objectives (e.g., questionnaire to many participants).

First of all, quantifying how much the user can keep their natural movements is highly controversial.
In our case, it is represented as the average (or sum) of the EMG signals.
Although the EMG can suggest the user's burden, it is not suitable for detecting motion's trajectory, which can be observed by an encoder on a knee joint.
Besides, other operators for the EMG signals can represent the user's burden from another perspective; for instance, one to extract the maximum value from a trial would indicate momentary burden. A threshold would suggest a range of burdens the user does not care that.

\subsection{Further improvement of sample-efficiency}
\label{subsec:discuss_efficiency}

The task defined in our framework is discretely given.
Therefore, the slope angle and the gait speed are discretized and assumed to be independent before optimizing the experiments.
However, as you know, they are continuous configurations, and the optimal gear ratio for them would be continuously changed.
Such over assumption would prevent maximization of the sample-efficiency.
A naive idea to resolve this is that the continuous configurations are set to the Gaussian process inputs in Bayesian optimization, $x$.
The proposed framework did not use any prior knowledge about the tasks, but we should actively use our prior knowledge to improve the sample-efficiency.
For example, dynamics would be useful information for walking on the different slope angles; and the lower burden in total should be desired for the elderly.

As another limitation attributed to the kernel method, the multi-task Bayesian optimization uses unnecessary samples for the target task, which wastes computational cost.
Specifically, the multi-task Bayesian optimization takes $\mathcal{O}(n^3)$ with $n$ samples, including ones from uncorrelated tasks. It might be reduced by integrating sparse Gaussian process~\cite{snelson2006sparse}.

\subsection{Combination with optimization of duty cycle}

The gear ratio is suitable for adjusting the generative power and the anti-torque, as mentioned in the section~\ref{subsec:principle}.
However, as hardware limitation, the gear ratio adjustment cannot ignore the settling time for control.
Our device takes few seconds to change the gear ratio, and therefore, it is difficult to frequently adapt the gear ratio to the user's motions in real-time.

To break this limitation, the combination with optimization of the duty cycle $D$ is one of the promising ways.
Since the duty cycle can be controlled electrically, the settling time is faster than one for the gear ratio adjustment.
Hence, the solution is considered that the gear ratio optimizes the main balance between the generative power and the anti-torque. Fine-tuning is conducted by the duty cycle during tasks in real-time.

\subsection{Optimal design of CVT}

In this paper, we have discussed the framework to optimize the gear ratio of CVT for multiple tasks.
On the other hand, however, the design of CVT still has room for discussion.
In particular, it is worth considering that our device is attached to the knee, which has a limited range of rotation (see Table~\ref{tab:params_mech}).
Although the current CVT is designed for infinite rotation, by optimally designing CVT under such prior knowledge about angular constraints, we may make the structure more compact, reduce energy loss, and increase the gear ratio range.

For example, if using a cone-type CVT like~\cite{singla2016optimization}, the half of the cone can transmit all the knee rotation.
Besides, its radius can probably be increased since mechanical interference is unlikely to occur, and the range of gear ratio can be expanded accordingly.

\section{Conclusion}
\label{sec:conclusion}

In this paper, we addressed the human-in-the-loop optimization framework for adjusting the gear ratio in CVT, installed into a biomechanical energy harvester.
With the CVT, the environment- and motion-specific (i.e., task-specific) optimal design of energy harvester can be regarded as the optimization problem of the gear ratio, which would be dependent on the tradeoff between harvesting energy and keeping the natural movements of the user.
However, this optimization must be terminated with a few trials since it is human-in-the-loop.
Therefore, our framework employed the multi-task Bayesian optimization to resolve this optimization problem sample-efficiently, which can reuse the data from previous tasks to the optimization for new tasks according to the correlations between them.

We conducted real-world experiments with the CVT-equipped energy harvester attached to two participants.
We found that the correlations between tasks improved the sample-efficiency and were adequately revealed in our framework.
As a result, experimental results suggested that our framework could efficiently optimize the task-specific gear ratios compared to the random search.
Specifically, our framework took around 20~minutes to optimize each task, although the random search took around 60~minutes.

As future work, a more appropriate optimization problem, i.e., the design of the score function, is essential for the practical use of our energy harvester.
We will then conduct experiments with a larger number of participants to analyze inter/extra participant variations and find a new score function suitable for human preferences.
Another direction would be to develop the harvest for other joints like back and shoulder joints with a similar principle.

\bibliographystyle{spmpsci}      
\bibliography{biblio}	

\end{document}